\pdfoutput=1

\documentclass[11pt]{article}

\usepackage[final]{acl}

\usepackage{times}
\usepackage{latexsym}
\usepackage{booktabs}
\usepackage{hyperref} 
\usepackage{tcolorbox}
\usepackage{subcaption}
\usepackage{array}
\usepackage{xcolor}
\usepackage{soul}
\usepackage{multicol,multirow}
\usepackage{todonotes}
\usepackage{xspace}
\definecolor{mygold}{HTML}{eeba0a}
\definecolor{mygrey}{HTML}{bac8ca}
\definecolor{mylightgreen}{HTML}{D1F694}
\definecolor{myblue}{HTML}{30C0F0}
\definecolor{myorange}{HTML}{F5AF22}
\definecolor{myred}{HTML}{FF576A}
\definecolor{darkred}{HTML}{91270F}
\definecolor{darkgreen}{HTML}{5E893E}
\definecolor{lightblue}{HTML}{b6ecfc}
\sethlcolor{mylightgreen}
\usepackage{makecell}
\usepackage{longtable}
\usepackage{amssymb}
\usepackage{pifont}
\definecolor{lightpink}{rgb}{1.0, 0.8, 0.9}

\usepackage[T1]{fontenc}

\usepackage[utf8]{inputenc}

\usepackage{microtype}

\usepackage{inconsolata}

\usepackage{graphicx}

\title{Protecting multimodal large language models against \\ misleading visualizations}

\author{
Jonathan Tonglet$^{1,2,3}$, Tinne Tuytelaars$^{2}$, Marie-Francine Moens$^{3}$, Iryna Gurevych$^{1}$
\\
        \textsuperscript{1} Ubiquitous Knowledge Processing Lab (UKP Lab), Department of Computer Science, \\ TU Darmstadt and National Research Center for Applied Cybersecurity ATHENE\\ 
\textsuperscript{2} Department of Electrical Engineering, KU Leuven\\
\textsuperscript{3} Department of Computer Science, KU Leuven\\
\href{https://www.ukp.tu-darmstadt.de}{www.ukp.tu-darmstadt.de}
}

\begin{document}
\maketitle
\begin{abstract}
Visualizations play a pivotal role in daily communication in an increasingly data-driven world. Research on multimodal large language models (MLLMs) for automated chart understanding has accelerated massively, with steady improvements on standard benchmarks. However, for MLLMs to be reliable, they must be robust to misleading visualizations, i.e., charts that distort the underlying data, leading readers to draw inaccurate conclusions. Here, we uncover an important vulnerability: MLLM question-answering (QA) accuracy on misleading visualizations drops on average to the level of the random baseline. To address this, we provide the first comparison of six inference-time methods to improve QA performance on misleading visualizations, without compromising accuracy on non-misleading ones. We find that two methods, table-based QA and redrawing the visualization, are effective, with improvements of up to 19.6 percentage points. We make our code and data available.\footnote{\href{https://github.com/UKPLab/acl2026-misleading-visualizations}{github.com/UKPLab/acl2026-misleading-visualizations}}

\end{abstract}

\section{Introduction}

\begin{figure*}[!ht]
\centering
\includegraphics[width=\textwidth]{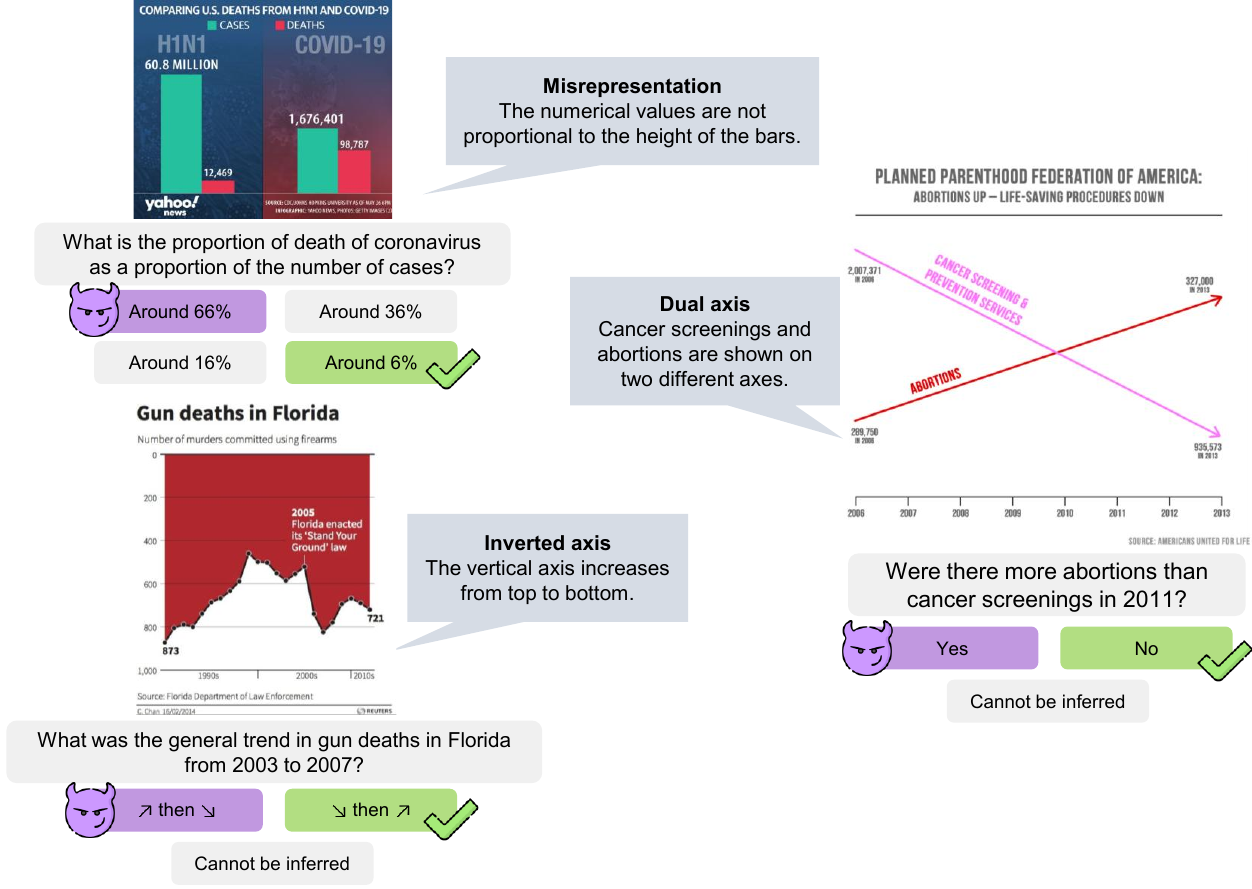}
\caption{Three examples of real-world misleading visualizations \citep{lo2022misinformed} with MCQs. The correct answer is colored in green, while the wrong answer supported by the misleader is colored in purple.}\label{fig4}
\end{figure*}

\begin{figure}[!ht]
\centering
\includegraphics[width=\linewidth]{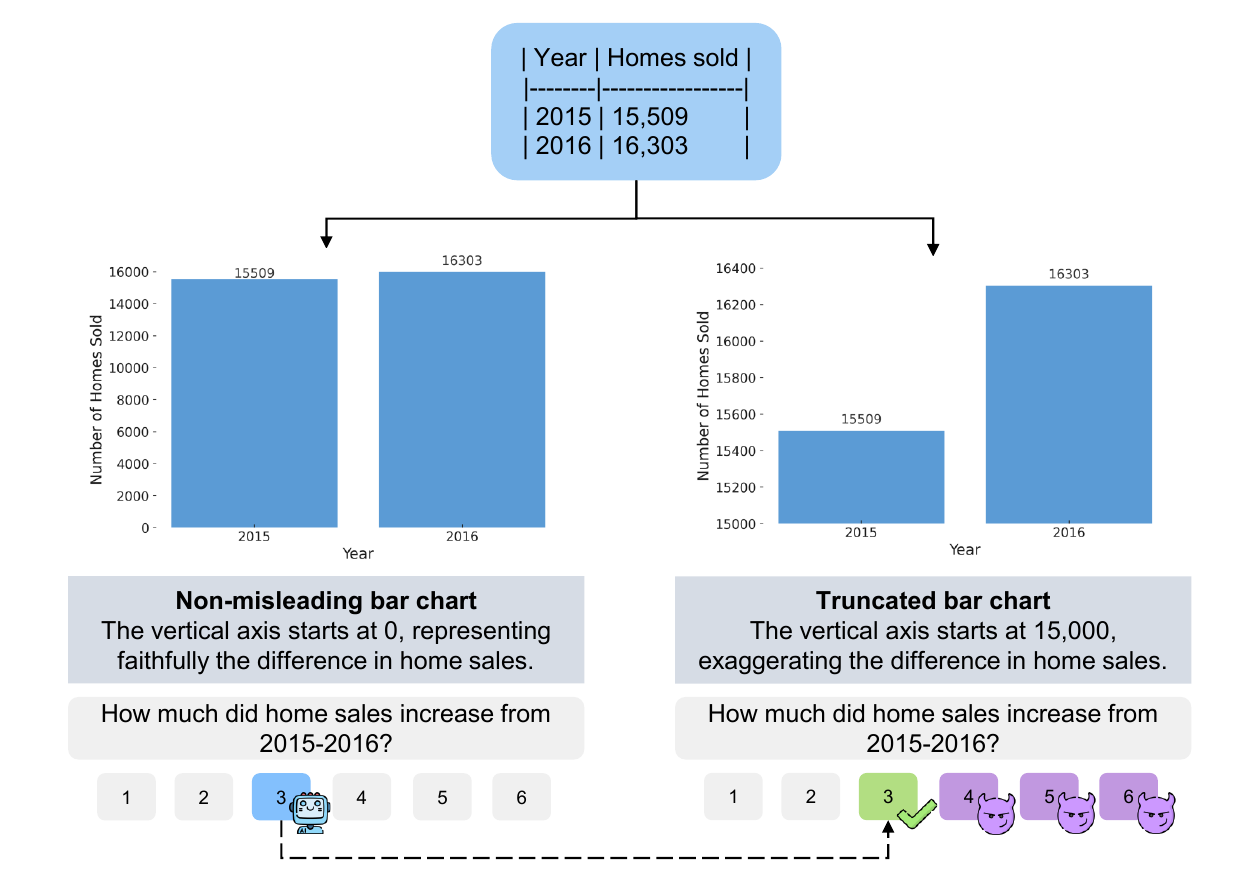}
\caption{Non-misleading and misleading visualizations of the same data \citep{10.1145/3380851.3416762} with a Likert-scale question where 1 means ``a little'' and 6 means ``a lot''. A consistent interpretation requires identical responses. However, the deceived reader chooses a higher value for the truncated bar chart.}\label{fig8}
\end{figure}

In an increasingly data-driven world, visualizations are widely used by scientists, journalists, governments, and companies to efficiently communicate data insights to a broad audience \citep{10787102}. They play a crucial role in crisis settings, such as during the COVID-19 pandemic, where charts were shared daily to inform the public \citep{10.1145/3411764.3445381}. In many cases, visualizations support a message more convincingly than showing the underlying data table directly to readers \citep{6876023}. However, visualizations can also be misleading \citep{2017-blackhatvis}. This occurs when design flaws, or misleaders, distort the underlying data in a way that prevents a correct interpretation by the reader \citep{tufte1983visual,10.1145/2702123.2702608,2017-blackhatvis,10.1145/3380851.3416762,10.1145/3313831.3376420,YANG2021298,lo2022misinformed,10.1145/3544548.3580910,10670488}. These misleaders appear across a wide range of visualizations, including bar and line charts, or choropleth maps. Recent taxonomies have documented over 70 types of misleaders observed in real-world examples \citep{lo2022misinformed,10670488}.  Major categories include axes manipulations, such as truncated and inverted axes, or the use of visual manipulations like 3D effects.
Figure \ref{fig4} shows three real-world examples of misleading visualizations \citep{lo2022misinformed} paired with multiple choice questions (MCQs). Given a question about the underlying data table, their misleaders lead readers to infer a misleading answer instead of the correct one. Figure \ref{fig8} illustrates how two visualizations of the same data, one non-misleading and the other misleading, can lead to different interpretations by readers.

Prior work has revealed the potential of misleading visualizations to deceive human readers. Some studies \citep{10.1145/2702123.2702608,10.1145/3544548.3581406,rho2023various,bharti2024chartom} have shown that such visualizations reduce human accuracy when answering MCQs about the underlying data. Other studies \citep{10.1145/2702123.2702608,10.1145/3233756.3233961,10.1145/3380851.3416762} have demonstrated that human readers tend to provide different  Likert-scale answers to a question depending on whether they view a misleading or non-misleading visualization of the same data table.

Misleading visualizations pose a serious threat to society. Due to their deceptive potential, they can be exploited by malicious actors to promote online disinformation \citep{2017-blackhatvis}. For instance, the charts in Figure \ref{fig4} mislead readers on sensitive topics such as COVID-19, abortion, and gun violence. Misleading visualizations can shift public opinion, even on polarized debates like Brexit \citep{10.1111/bjso.12787}.

Multimodal large language models (MLLMs) have sparked a massive interest in automated chart understanding research \citep{10787102},  with steady progress on reference benchmarks such as  ChartQA \citep{masry-etal-2022-chartqa,10.1111/cgf.14573}.
However, existing research has overlooked the important threat posed by misleading visualizations. If MLLMs, like humans, are easily deceived by misleading visualizations, malicious actors could exploit this vulnerability. MLLMs confronted with misleading visualizations could present incorrect interpretations of the underlying data to users, contributing to the spread of disinformation and reinforcing human belief in it. These risks underscore the urgency of thoroughly assessing the vulnerability of MLLMs to misleading visualizations and developing effective mitigation strategies.

In this work, we conduct the first comprehensive study to assess and mitigate the vulnerability of 19 MLLMs of varying sizes to misleading visualizations. In Experiment 1, we compare the question-answering (QA) accuracy of MLLMs on misleading and non-misleading visualizations. In Experiment 2, we evaluate the consistency of MLLMs, i.e., whether they provide the same response to a Likert-scale question when shown either a misleading or a non-misleading visualization of the same underlying data.
The results of both experiments demonstrate that MLLMs, like humans, are indeed vulnerable to misleading visualizations. 
In Experiment 3, we compare six inference-time correction methods to reduce MLLMs' vulnerability to the misleading visualizations from Experiment 1. 

In summary, our contributions are twofold:
(1) We present the first extensive analysis of the vulnerabilities of 19 MLLMs to 17 types of misleading visualizations, evaluating both QA accuracy and Likert-scale consistency.
(2) We propose the first analysis of six correction methods to mitigate the negative impact of misleaders on QA tasks.

\section{Methodology}

\subsection{Datasets}

\textbf{Experiments 1 and 3} We compare the question-answering (QA) accuracy of MLLMs across two datasets: (a) a misleading visualization dataset containing $n$ = 143 instances, and featuring 17 types of misleaders, defined in Appendix \ref{secA1} \citep{10.1145/3544548.3581406,rho2023various,bharti2024chartom,lo2022misinformed}; (b) a non-misleading visualization dataset ($n$ = 124) \citep{10.1145/3544548.3581406,rho2023various,bharti2024chartom,7539634}, also combining real-world and synthetic cases; and (c) ChartQA \citep{masry-etal-2022-chartqa}, the reference real-world non-misleading benchmark ($n$ = 2500).   Datasets (a) and (b) combine three existing resources and one introduced in this work. First, CALVI \citep{10.1145/3544548.3581406} includes 45 misleading and 15 non-misleading visualizations based on synthetic data, each paired with an MCQ. CALVI was initially designed and is the reference resource for evaluating humans.
Second, CHARTOM \citep{rho2023various,bharti2024chartom} contains 56 samples, including 28 MCQs, 20 free-text questions, and 8 rank questions. Each question is linked to two visualizations, one misleading and one non-misleading. Like CALVI, the underlying data is synthetic, and the test was originally designed to evaluate humans.
Third, VLAT \citep{7539634}, the reference dataset to evaluate humans on non-misleading cases, provides  12 visualizations, each paired with three to seven MCQs, for a total of 53 instances. The visualizations are based on real-world data.
Fourth, we introduce a dataset of 42 real-world misleading visualizations, each annotated with an MCQ with three to four choices. They come from a collection annotated with the misleaders affecting them \citep{lo2022misinformed}, which inspired the synthetic examples in CALVI. We manually create MCQs, using CALVI and CHARTOM as references. By incorporating visualizations and questions about real-world data, we introduce direct conflicts with MLLMs' parametric knowledge, allowing us to assess their vulnerability in real-world scenarios.

\textbf{Experiment 2} We use four pairs of visualizations,  one misleading and the other non-misleading, designed by \citet{10.1145/3380851.3416762}. 

Table \ref{tab4} provides detailed dataset statistics.

\begin{table}
\resizebox{\linewidth}{!}{
\begin{tabular}{@{}lllll@{}}
\toprule
Dataset & Experiments & $n$  & $m$ & Question types \\
\midrule
\textbf{Misleading visualizations}        & 1 \& 3 & 143  &17 & MCQ, Free-text, rank \\
$\hookrightarrow$ CALVI (misleading) & 1 \& 3 &  45 & 14 &  MCQ   \\
$\hookrightarrow$ CHARTOM (misleading) & 1 \& 3 & 56  & 7 & MCQ, Free-text, rank   \\
$\hookrightarrow$ Real-world & 1 \& 3 & 42 &  12 & MCQ   \\
\midrule
\textbf{Non-misleading visualizations}       & 1 \& 3 & 124 & - & MCQ, Free-text, rank   \\
$\hookrightarrow$ CALVI (non-misleading)  & 1 \& 3 & 15 & - & MCQ \\
$\hookrightarrow$ CHARTOM (non-misleading) & 1 \& 3 & 56  & - & MCQ, Free-text, rank \\
$\hookrightarrow$ VLAT  & 1 \& 3 &  53 & - & MCQ  \\
\midrule
ChartQA & 1 &  2500 & - & Free-text  \\
\midrule
Lauer \& O'Brien & 2 & 8  & 3 &  Likert-scale \\
\bottomrule
\end{tabular}}
\caption{Datasets statistics. $n$ is the number of instances. $m$ is the number of misleader categories.}\label{tab4}
\end{table}

\subsection{Correction methods}

\begin{figure*}
\centering
\includegraphics[width=0.8\textwidth]{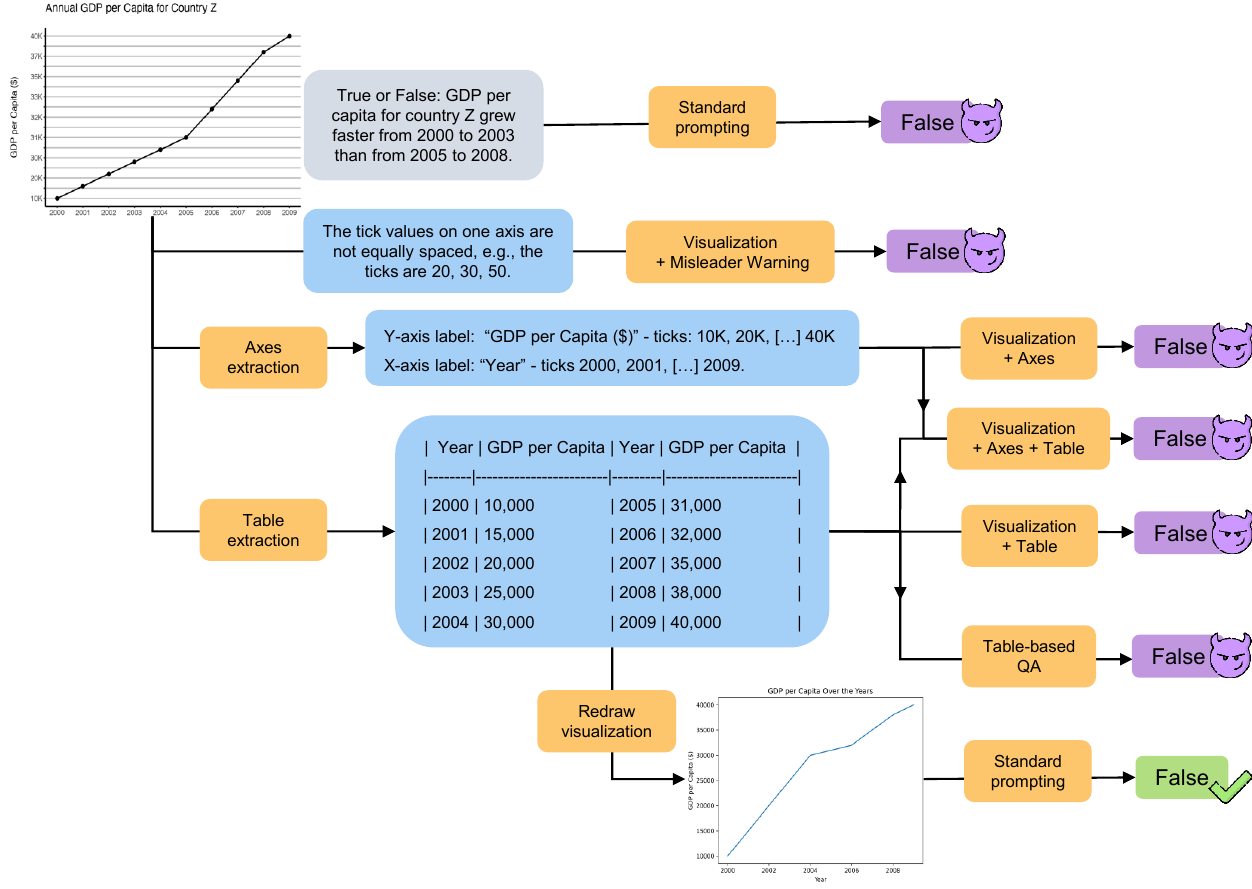}
\caption{Illustration of the six inference-time correction methods applied to a misleading visualization from CALVI \citep{10.1145/3544548.3581406}. The visualization suffers from inconsistent tick intervals on the y-axis.}\label{fig3}
\end{figure*}

In Experiment 3, we compare six inference-time correction methods, illustrated in Figure \ref{fig3}, to increase QA accuracy on misleading visualization without compromising it on non-misleading ones.

\textbf{(1) Misleader warning}: we insert in the prompt a message to make the MLLM aware of the misleader in the visualization. The message is identical for all instances with the same misleader.

\textbf{Providing (2) the axes data, (3) the data table, (4) or both}: we prompt the MLLM in zero-shot to extract the axes labels or underlying data table. The axes and tables are formatted as text strings. We do not impose formatting constraints on the axes and tables. The axes, the table, or both are provided as additional prompt input.

\textbf{(5) Table-based QA}: after extracting the table with the MLLM, we provide it alone to a text-only LLM, reframing the task as table-based QA \citep{liu-etal-2023-deplot,liu-etal-2023-matcha,kim-etal-2025-simplot}. 

\textbf{ (6) Redrawn visualization}: after extracting the table with the MLLM, we provide it to a text-only LLM, which generates Python code to create a visualization using Matplotlib \citep{Hunter:2007}. If the code compiles successfully, the generated visualization replaces the original one in the QA prompt; otherwise, the original one is used.

\subsection{Implementation details}

\paragraph{Metrics} The evaluation of QA accuracy depends on the question type. For MCQs and rank questions, we evaluate the exact match. For free-text questions, which all expect numerical answers, we use a relaxed accuracy with a 5\% tolerance threshold, following the standard ChartQA setup \citep{masry-etal-2022-chartqa}. For ChartQA, we report the scores established in prior benchmark studies \citep{chen2024internvl,lu2024ovisstructuralembeddingalignment}.

\paragraph{Models} We conduct experiments with 19 MLLMs released in 2023-2024 on a machine with two A100 GPUs, including 11 open-weight MLLMs from the Llava-Next \citep{10655294}, Qwen2VL \citep{wang2024qwen2}, Ovis1.6 \citep{lu2024ovisstructuralembeddingalignment}, and InternVL2.5 \citep{chen2024expanding} families. We also include five commercial models, GPT4 and GPT4o \citep{openai2023gpt4}, Gemini-1.5-flash and -pro \citep{google2024gemini}, and Claude-3.5-sonnet \citep{claude2024}, as well as three open-weight MLLMs specialized in chart understanding: Llava-Chart-Instruct \citep{10670526}, TinyChart (using the Direct approach) \citep{zhang-etal-2024-tinychart},  and ChartGemma \citep{masry-etal-2025-chartgemma}. Qwen2.5-7B \citep{yang2024qwen2} serves as the LLM for table-based QA and visualization redrawing. We use the transformers library \citep{wolf-etal-2020-transformers} to access all open-weight models.   All prompts are provided in Appendix \ref{sec:prompts}. We set the temperature to 0. Following the standard ChartQA evaluation setup, all (M)LLMs are prompted in zero-shot.

\paragraph{Random baseline} We compare the MLLMs against a random baseline. Its accuracy is 1/n for an MCQ with n choices, 0 for free-text,  and 1/n! for ranking with n items.

\section{Results}\label{sec2}

\subsection{Experiment 1 - Assessing MLLM vulnerabilities (Accuracy)}

\begin{figure*}[!ht]
\centering
\includegraphics[width=\textwidth]{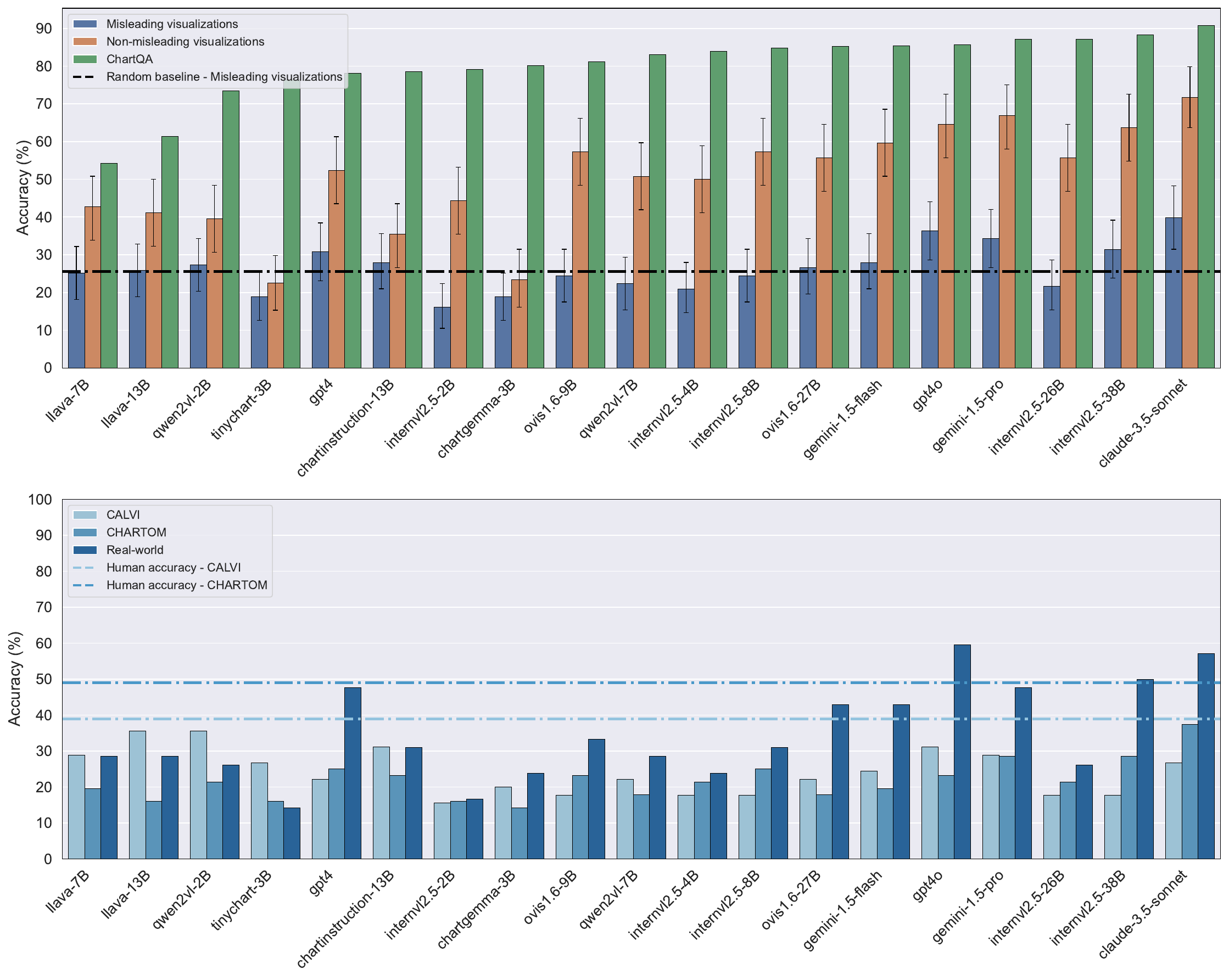}
\caption{Top: QA accuracy (\%) on the misleading visualization, non-misleading visualization, and ChartQA datasets. Confidence intervals are reported for the misleading and non-misleading datasets. The horizontal dashed line indicates the accuracy of the random baseline on misleading visualizations. Models are sorted by increasing accuracy on ChartQA. Bottom: Accuracy (\%) of various MLLMs on subsets of the misleading visualizations. The horizontal dashed lines indicate average human accuracy on CALVI and CHARTOM.}\label{fig1}
\end{figure*}

Experiment 1 assesses the ability of MLLMs to answer questions about the underlying data given a visualization.
The upper part of Figure \ref{fig1} presents the QA accuracy of 19 MLLMs across the three datasets, sorted by increasing accuracy on ChartQA. We report bootstrapped (n=5000) confidence intervals (CIs) for the misleading and non-misleading datasets. The bottom part displays the accuracy for the three subsets constituting the misleading visualizations dataset. Additional results are provided in Appendix \ref{sec:additional_results}. Experiment 1 reveals three key findings.

\textbf{Performance is worse on misleading visualizations.} All MLLMs perform substantially worse on misleading visualization compared to the non-misleading visualizations dataset, with accuracy dropping by up to 34.8 percentage points (pp), with an average of 23.8 pp. The decline is even more pronounced compared to ChartQA, reaching up to 65.5 pp, with an average of  53.8 pp. Furthermore, the mean MLLM accuracy on misleading visualizations (26.4\%) is comparable to the random baseline score (25.6\%). These results demonstrate that MLLMs struggle to interpret data distorted by misleaders correctly. 

For comparison, average human accuracy reported on CALVI and CHARTOM is 80 and 89\% for non-misleading visualizations, dropping to 39 and 49\% for misleading ones, respectively \citep{10.1145/3544548.3581406,rho2023various,bharti2024chartom}.  On the same datasets, MLLMs achieve an average of 52 and 48\% on non-misleading visualizations and 24 and 22\% on misleading ones, falling short of human performance in all cases.

The ranking of MLLMs by accuracy varies a lot across the three datasets. The two best-performing models on the CALVI subset of misleading visualizations rank among the worst on ChartQA, placing second and third from the bottom. This suggests that achieving high accuracy on misleading visualizations is unlikely to emerge naturally from improvements on ChartQA, underscoring the need for dedicated mitigation methods.

\textbf{Performance is higher on real-world misleading visualizations than on other subsets.} MLLMs perform better on the real-world subset, achieving an average accuracy of 34.7\%, compared to 24\% on CALVI and 22\% on CHARTOM.  In general, the best-performing MLLMs on misleading visualizations, Claude-3.5-sonnet, GPT4o, and Gemini-1.5-pro, primarily distinguish themselves through their higher performance on this subset. We hypothesize that this advantage stems from their parametric knowledge, which extends beyond 2022, the endpoint of the real-world subset. This allows these models to answer some questions based on stored knowledge of past events, without relying on the visualization's content. Appendix \ref{sec:real-world-knowledge} discusses further the impact of parametric knowledge. Other factors, which are not quantified in this work, might explain the better performance on the real-world subset, such as the chart aesthetics and visual complexity, or the difficulty of the MCQs compared to CALVI and CHARTOM.

\textbf{Performance varies by type of misleader.} Not all misleading visualizations are equally deceptive. This was already observed with human readers \citep{10.1145/3544548.3581406,rho2023various,bharti2024chartom}, and is here further confirmed with MLLMs. The following analysis considers only misleaders with at least five occurrences in the dataset. The two misleaders on which MLLMs perform best are \textit{misleading annotations} and \textit{misrepresentation}, with average accuracies of 50.5 and 44.4\%. The two on which they perform worse are \textit{area encoding} and \textit{cherry-picking} at 7.4 and 10.5\%. For the two most represented misleaders, \textit{truncated axis} and \textit{inverted axis}, the MLLMs achieve average accuracies of 25.1 and 28.1 \%. There are notable differences in which misleaders pose greater challenges to MLLMs compared to humans, and vice versa. For instance, \textit{area encoding} is highly misleading for MLLMs, but not for human readers, who achieve an average accuracy of 91.5\% \citep{rho2023various}. Conversely, both humans and MLLMs struggle with misleaders such as \textit{3D effects} \citep{10.1145/3544548.3581406}. The use of a \textit{dual axis} is a misleader that poses greater difficulty for humans, who score an average accuracy of 16.1\% on CHARTOM \citep{rho2023various}, compared to MLLMs, which achieve an average of 35.5\%. These results highlight a notable difference in how misleaders deceive humans and MLLMs. This suggests that the methods needed to make MLLMs more robust to misleading instances will, at least in part, differ from those designed to protect human readers \citep{10.1145/3491102.3502138}.

\subsection{Experiment 2 - Assessing MLLM vulnerabilities (Consistency)}

Experiment 2 evaluates whether MLLMs interpret consistently two visualizations that represent the same underlying data table, one misleading and the other non-misleading. This is assessed using a  Likert-scale question, such as in Figure \ref{fig8}. Unlike in Experiment 1, there is no ground truth answer. Instead, since both visualizations depict the same underlying data, a model is considered misled if it assigns different scores to the two. We assess the significance of the difference in ratings with a Wilcoxon signed-rank test (p-value (p) $\leq$ 0.05).

\begin{figure}
\centering
\includegraphics[width=\linewidth]{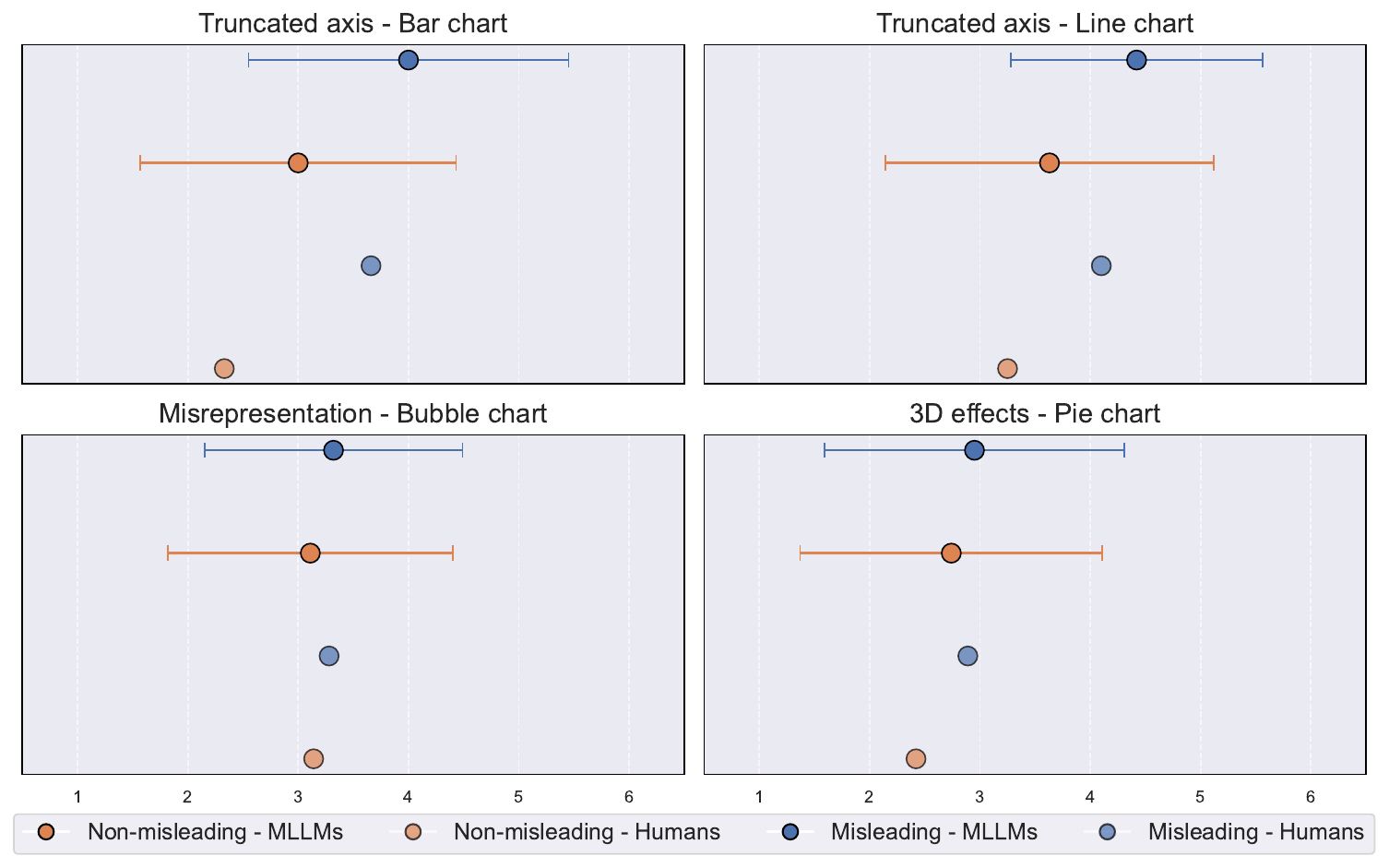}
\caption{Average Likert-scale ratings (1 to 6) in Experiment 2. Average MLLM results are reported with standard deviations. Average human results are reported from \citet{10.1145/3380851.3416762}.}\label{fig6}
\end{figure}

Figure \ref{fig6} summarizes the average Likert-scale ratings from all 19 MLLMs for each visualization pair. Average human results from prior work \citep{10.1145/3380851.3416762} are included for comparison. The average rating difference between the misleading and non-misleading visualizations is similar for humans and MLLMs, though MLLMs tend to assign slightly higher scores overall. For MLLMs, the inconsistency in ratings is significant for the two visualization pairs involving \textit{truncated axis}: one with bar charts (p = 7e-3) and one with line charts (p = 6e-3). For the other two pairs, \textit{misrepresentation} with bubble charts (p = 0.36) and \textit{3D effects} with pie charts (p = 0.49), the differences are not significant. These results give an initial indication that MLLMs are unable to consistently recognize the same underlying data table behind a misleading and a non-misleading version of the same visualization, particularly for \textit{truncated axis}. However, these results are only preliminary and need to be further validated in future work with a larger set of MLLMs on more data pairs.

\subsection{Experiment 3 - Correction methods}

\begin{figure*}
\centering
\includegraphics[width=\textwidth]{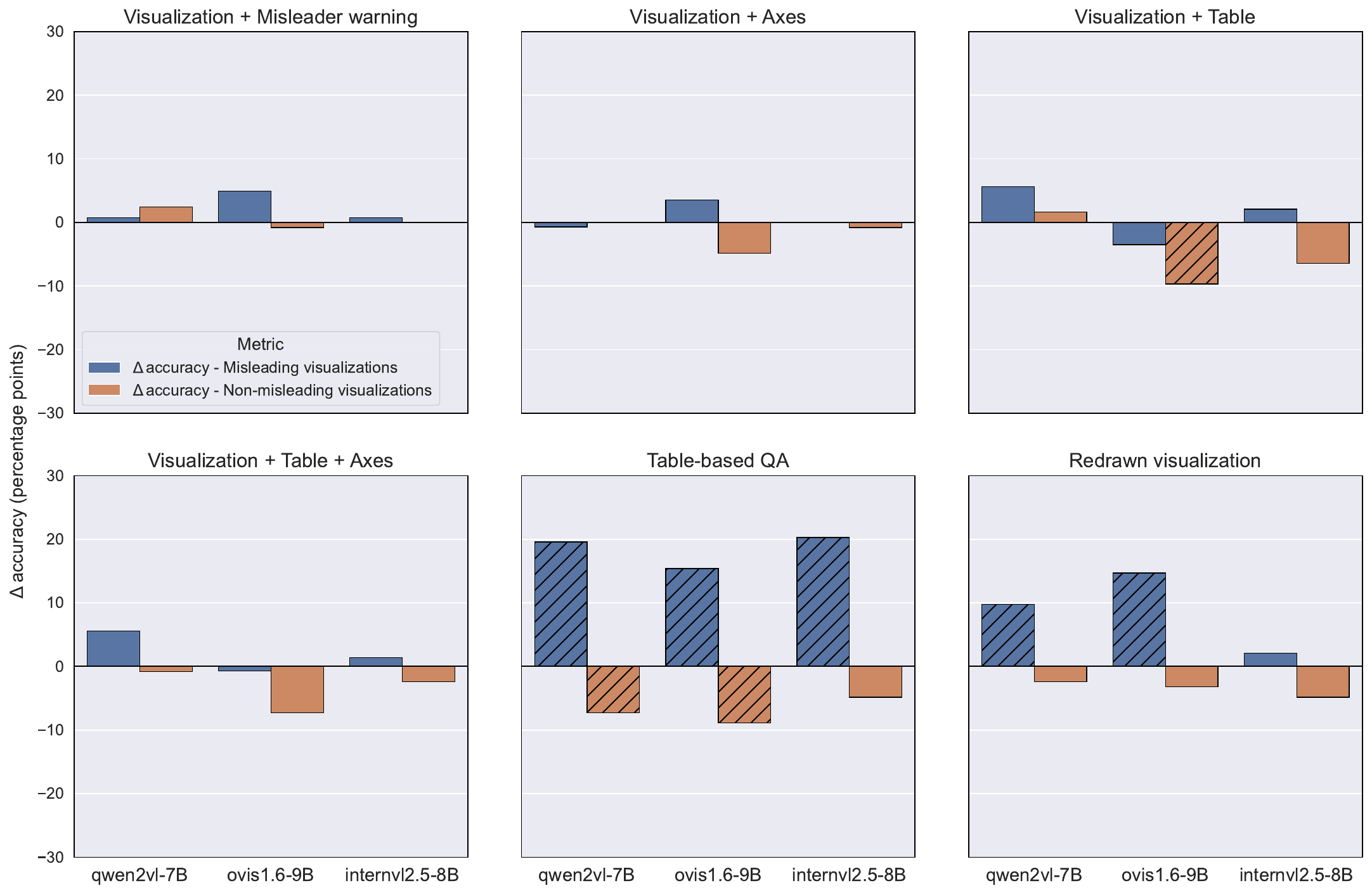}
\caption{Change ($\Delta$) in accuracy (percentage points) compared to Experiment 1 using different inference-time correction methods. Statistically significant changes (p$\leq$0.05) are hashed.}\label{fig2}
\end{figure*}

Having established that MLLMs are vulnerable to misleading visualizations, we examine six correction methods designed to enhance their robustness. Our focus is on increasing accuracy on the misleading visualizations dataset from Experiment 1, while maintaining performance on the non-misleading dataset. We evaluate the impact of six correction methods using three open-weight, mid-sized MLLMs, Qwen2VL-7B, Ovis1.6-9B, and InternVL2.5-8B, all of which performed below the random baseline in Experiment 1. Appendix \ref{sec:additional_results_exp3} provides the detailed results. Statistical significance of the changes in accuracy is assessed using the two-sided McNemar test (p $\leq$ 0.05) \citep{McNemar_1947}. Figure \ref{fig2} shows the change in accuracy relative to Experiment 1 across the six methods: (1) inclusion of a warning message, (2) inclusion of extracted axes, (3) inclusion of the extracted table, (4) inclusion of both axes and table, (5) table-based QA, and (6) redrawing the visualization.

\textbf{Two correction methods are effective: table-based QA and redrawing the visualization.} The most effective approach to counter misleaders is table-based QA, which yields significant improvements on misleading visualizations, ranging from 15.4 to 19.6 pp. Originally proposed as a general approach for chart understanding \citep{liu-etal-2023-deplot,liu-etal-2023-matcha}, table-based QA further shows its effectiveness for countering misleading charts. Its strength lies in replacing the visualization with a data table, thereby eliminating misleaders that do not transfer to the tabular format, such as \textit{inverted} and \textit{truncated axis}. However, this method proves less effective against misleaders that persist even when the data is presented in tabular format, such as \textit{inappropriate item order}. 
Table-based QA incurs a significant cost for two models on non-misleading visualizations, by up to 8.5 pp. This degradation is due to errors in the intermediate table extraction step, which introduce incorrect values or cause relevant information to be lost. Appendix \ref{sec:additional_results_exp3} reports table-based QA performance using DePlot \citep{liu-etal-2023-deplot} and MatCha \citep{liu-etal-2023-matcha}, two methods developed specifically for table extraction.

Another promising correction method is redrawing the visualization. This method aligns with prior efforts to make human readers more robust to misleading visualizations \citep{10.1145/3491102.3502138}. It leads to significant, though more modest, improvements for two MLLMs. The visualization is redrawn using the default settings of the Python library Matplotlib \citep{Hunter:2007}, which inherently avoids certain misleaders like inverting the axes or using inconsistent tick intervals. In Figure \ref{fig3}, the redrawn visualization corrects the misleading trend line by eliminating inconsistent tick intervals, leading to a more accurate representation of the data. This correction method is effective only if the generated code compiles successfully on a Python interpreter; otherwise, the original visualization is used. The lowest redrawing success rates are observed for scatter plots (79\%) and stacked bar charts (80\%), likely due to their higher visual complexity and number of elements, which require Qwen2.5-7B to produce longer Python code. Unlike table-based QA, there are no significant decreases in accuracy on non-misleading visualizations, making it a more attractive option when preserving performance on non-misleading visualizations is a priority.

Appendix \ref{sec:additional_results_exp3} discusses the impact of combining these two effective methods with Chain-of-Thought prompting \citep{10.5555/3600270.3602070}.

\textbf{Table extraction is a key intermediate step.} Both table-based QA and redrawing the visualization depend heavily on the accuracy of the intermediate table extraction step. This extraction step is often non-trivial, particularly for maps or scatter plots that contain many visual elements. Additionally, misleaders themselves can degrade MLLMs’ ability to extract data accurately. This is particularly prevalent for \textit{inverted axis}, \textit{3D effects}, and \textit{dual axis}.
Perfect accuracy in table extraction is not always necessary; its importance depends on the nature of the question. Some questions can be answered correctly if the extracted table preserves the overall trend of the underlying data, while others require exact values. We conduct a qualitative analysis of table extraction in Appendix \ref{secA3}.

\textbf{Other correction methods are not effective.} None of the remaining correction methods yields significant improvements. While using the extracted table alone is the most effective method, combining it with the misleading visualization does not yield similar gains. The MLLMs are biased toward using the image rather than the table. 

Although many misleaders rely on axes manipulations, providing the extracted axes has no significant effect on performance. A manual error analysis showed that the axes extraction step is more often accurate than table extraction, as reported in Appendix \ref{sec:axis_extraction}. However, unlike the table, the axes alone are insufficient to answer the question. The MLLM still needs to combine the visualization image with cues from the axes indicating the presence of misleaders. The results show that this combination is challenging. This further suggests that removing or modifying the misleading visualization, rather than merely supplementing it with additional prompt input, is a key factor in the success of a correction method.   

Adding a warning message produces only non-significant changes, with accuracy gains of at most 4.9 pp, mostly limited to the real-world subset. This warning-based approach assumes prior knowledge of the specific misleader in the visualization, making the reported results an upper bound on its effectiveness. In practice, a classifier needs to detect first the presence of misleaders, which remains a challenging task \citep{10679256,10771149}. Given the already low results obtained with ground-truth misleader labels, this correction method appears unpromising overall. However, training a highly accurate misleader detection model could enable the selective application of correction methods. This would offer two key advantages: (1) avoiding the application of correction methods to non-misleading visualizations, thereby eliminating any risk of negative impact; and (2) allowing the selection of the most suitable correction method based on the type of misleader.

\section{Related work}

The correction methods, particularly misleader warnings and redrawing the visualization, relate to prior work on visualization linters \citep{10.1111/cgf.13975,9552878,10.1145/3491102.3502138}: rule-based methods to detect and correct misleaders. While these tools proved useful to make human readers more robust to misleading visualizations \citep{10.1145/3491102.3502138}, they can only operate with specific chart design tools \citep{Hunter:2007,10.1111/cgf.13975,9552878} or a very limited set of chart and misleader types \citep{10.1145/3491102.3502138}. In contrast, the correction methods we explored do not impose such restrictions. Table-based QA was originally introduced as a general approach for chart understanding by \citet{liu-etal-2023-deplot,liu-etal-2023-matcha}. Although this approach has been surpassed on standard benchmarks by more recent methods that treat visualizations purely as images \cite{10670526, masry-etal-2025-chartgemma}, our results reveal that table-based QA finds a new purpose as an effective method for counteracting misleading visualizations.

Our findings are further supported by prior work \citep{10670574} and parallel studies \citep{10.1111/cgf.70137,chen-etal-2025-unmasking,valentim2025plotthickensquantitativepartbypart}, which also report MLLM vulnerabilities to misleaders \citep{10670574,10.1111/cgf.70137,chen-etal-2025-unmasking} and to other design decisions like the color palette \citep{valentim2025plotthickensquantitativepartbypart}. Our study differentiates itself in several key ways. Prior work \citep{10670574} only evaluated GPT4 on six misleading visualizations, providing initial hints but lacking depth to fully establish the presence of a vulnerability. Unlike parallel works, we include real-world visualizations and show that the vulnerability of MLLMs is lower in these cases. Furthermore, we provide the first evaluation of six correction methods. Crucially, we examine the trade-off between improving QA performance on misleading visualizations and preserving accuracy on non-misleading ones, an aspect not addressed in other works.

\section{Conclusion}\label{sec3}

Our findings highlight the vulnerability of MLLMs to misleading visualizations.  Identifying this vulnerability fills a critical gap in the research on automated chart understanding. While MLLMs achieve strong performance on standard benchmarks such as ChartQA, they remain vulnerable to misleading visualizations. This raises serious concerns about their reliability in real-world settings, especially given the potential for such vulnerabilities to be exploited by malicious actors to spread disinformation \citep{2017-blackhatvis}. To mitigate this issue, we evaluated six inference-time correction methods, two of which, table-based QA and redrawing the visualization, demonstrated significant improvements. However, correction methods often come at the cost of reduced accuracy on non-misleading visualizations.

\section*{Limitations}

We identify three limitations to our work. 

First, the visualization redrawing method does not support maps, as the Matplotlib library \citep{Hunter:2007} lacks sufficient functionality for rendering high-quality maps. 

Second, we assume prior knowledge of the chart type (e.g., bar, line) to generate prompts for axes extraction and redrawing. This is a reasonable assumption, as the chart type can either be provided by a human user or accurately predicted by a classifier as a preprocessing step. 

Third, existing misleading visualization datasets are small compared to standard datasets in the automated chart understanding literature, such as ChartQA. Furthermore, they do not provide an equal representation of all misleaders, with \textit{truncated} and \textit{inverted axes} being the two most represented categories. This imbalance is prevalent in CALVI \citep{10.1145/3544548.3581406} and, to a lesser extent, in real-world data. Despite these limitations, CALVI is the reference dataset for assessing human vulnerability to misleading visualizations. It was carefully curated by experts in data visualization and validated on a large sample of human subjects. Therefore, it constitutes a valid resource for evaluating MLLMs. The properties of CALVI are similar to those of VLAT \citep{7539634}, an expert-created dataset for non-misleading chart understanding, which acts as a reference despite being much smaller than datasets such as ChartQA. 
Furthermore, the real-world distribution of misleaders is inherently imbalanced \citep{lo2022misinformed}, with categories such as \textit{truncated axis} being more prevalent than others. Hence, scaling the real-world dataset would not address the imbalance issues of synthetic datasets, such as CALVI.

\section*{Ethics statement}

\paragraph{Social impact} Misleading visualizations are a prevalent form of multimodal misinformation. Our results show the importance of protecting not only humans but also MLLMs from misleading visualizations. Our correction methods provide the first promising results in that direction.

\paragraph{Risks} While the proposed correction methods improve the performance of MLLMs on misleading visualizations, they do not guarantee a correct answer. In general, human users should not blindly trust the output of chart understanding systems.

\paragraph{Dataset access} We release the CALVI, VLAT, Likert-scale, and real-world datasets under a CC-BY-SA-4.0 license. The real-world dataset is intended solely for research purposes. CHARTOM can be accessed by contacting its authors \citep{bharti2024chartom}.

\paragraph{AI assistants use}  AI assistants were used in this work to assist with writing by correcting grammar mistakes and typos.

\section*{Acknowledgments}

This work has been funded by the LOEWE
initiative (Hesse, Germany) within the
emergenCITY center (Grant Number:
LOEWE/1/12/519/03/05.001(0016)/72),
by the German Federal Ministry of Education
and Research and the Hessian Ministry of Higher
Education, Research, Science and the Arts within
their joint support of the National Research
Center for Applied Cybersecurity ATHENE, and by the Flanders AI Research Program. We
gratefully acknowledge the support of Microsoft
with a grant for access to OpenAI GPT models via
the Azure cloud (Accelerate Foundation Model
Academic Research). The figures have been designed using resources from Flaticon.com. We want to express our gratitude to Niklas Traser for conducting an initial exploration of the real-world data, to Jan Zimny for our insightful discussions on misleading visualizations, and to Germàn Ortiz, Manisha Venkat, and Max Glockner for their feedback on a draft of this work.

\bibliography{anthology,custom}

\newpage

\appendix

\section{Definition of misleaders}\label{secA1}

Table \ref{tab1} provides the definitions of all the misleaders considered in this work \citep{lo2022misinformed,10.1145/3544548.3581406}, with their number of occurrences.

\begin{table*}
\resizebox{\linewidth}{!}{
\begin{tabular}{@{}ll@{}}
\toprule
Misleader &  Definition  \\
\midrule
Inverted axis ($n$=26) &    An axis is oriented in an unconventional 
direction   \\
& and the perception of the data is reversed \citep{lo2022misinformed}. \\
Truncated axis ($n$=21) &      The axis does not start from zero or is truncated   \\
&  in the middle, resulting  in an exaggerated difference \\
& between the two bars \citep{lo2022misinformed}.  \\
Inappropriate axis range ($n$=15) &   The axis range is either too broad or too
narrow to   \\ 
& accurately   visualize the data, allowing changes to be \\
&   minimized or maximized depending on \\
& the author’s intention \citep{lo2022misinformed}.\\ 
Inconsistent tick intervals ($n$=12) &   Cases with varying intervals between the ticks \citep{lo2022misinformed}.\\
3D effects ($n$=12) &  The closer something is, the larger it appears, \\
&  despite being the same size
in 3D perspective \citep{lo2022misinformed}. \\
Inappropriate item order ($n$=9) &  The axis labels or legends appear to be in a random \\ 
& order  due to manipulation of data ordering \citep{10.1145/3544548.3581406}.\\
Inappropriate aggregation  ($n$=8)  &     Aggregating data in an improper way that leads to   \\
&  inaccurate conclusions \citep{10.1145/3544548.3581406}.  \\
Dual axis ($n$=8) &   Two independent axes are layered on top of \\ 
&  each other with inappropriate scaling \citep{lo2022misinformed}.\\ 
Misrepresentation ($n$=7) &   The value labels provided do not match \\
& the visual
encoding \citep{lo2022misinformed}.\\ 
Cherry picking ($n$=6) &  Selecting only a subset of data to display, which  \\ 
&  can be misleading  if one is asked to infer something \\
&  about the whole set of data \citep{10.1145/3544548.3581406}. \\
Misleading annotations ($n$=5) &  Annotations that contradict or make it harder \\ 
&  to read  the visualization \citep{10.1145/3544548.3581406}. \\
Area encoding ($n$=5) &  Linearly encoding the values  as areas    \\
& leads the readers to consistently  \\
& underestimate the values \citep{lo2022misinformed}. \\
Concealed uncertainty ($n$=3) &   Not displaying uncertainty in visualizations may   \\
& misrepresent the certainty  in  the underlying data.  \\
&  In the case of  prediction  making, this can misguide   \\ 
& the viewers to falsely overconfident conclusions \citep{10.1145/3544548.3581406}. \\
Missing normalization ($n$=3) &  Displaying unnormalized data in absolute quantity  \\
&  when normalized  data in relative quantity  \\
& is of interest \citep{10.1145/3544548.3581406}.\\
Inappropriate use of pie chart ($n$=1) &  When a pie chart is used for non-part-to-whole data, \\
&  it creates confusion for the
audience, who may \\
&  misinterpret the significance of a given section \citep{lo2022misinformed}.\\
Missing data ($n$=1) &  A visual representation implies data exist but  \\
& the data is actually missing \citep{10.1145/3544548.3581406}.\\
Overplotting ($n$=1) &  Displaying too many things on a plot can \\
& obscure parts of the data \citep{10.1145/3544548.3581406}.\\

\bottomrule
\end{tabular}}
\caption{The misleaders included in this work, with their number of occurrences ($n$).}\label{tab1}
\end{table*}

\section{Prompts}\label{sec:prompts}

Figures \ref{fig:axes-extraction-prompt}, \ref{fig:table-extraction-prompt}, and \ref{fig:redrawing-prompt} provide the prompt for the intermediate metadata extraction and visualization redrawing steps. Figure \ref{fig:qa-prompt} provides the QA prompt variants. Different parts of the prompt are included depending on the question type and additional input of correction methods. 

\begin{figure}
\centering
\begin{tcolorbox}[colback=gray!5, colframe=gray!80, title=Table extraction prompt, width=\linewidth]
Generate the underlying data table of the figure below. Change columns with |, change row by starting a new line. Provide only the table as output.
\end{tcolorbox}
    \caption{Table extraction prompt.}
\label{fig:table-extraction-prompt}
\end{figure}

\begin{figure}
\centering
\begin{tcolorbox}[colback=gray!5, colframe=gray!80, title=Axes extraction prompt, width=\linewidth]
\textbf{For maps}: What is the legend and its categories (with their colors) in this map? Answer only with the content of the legend and its categories (with their colors).
\\

\textbf{For pie charts}:  What are the categories in this pie chart? Answer only with the categories.
\\

\textbf{Other chart types}: What are the axis labels and ticks of this chart? Answer only with the axis labels and the tick values, going from the bottom-left to the top-left corner of the chart for the y-axis and from the bottom-left to the bottom-right corner for the x-axis.
\end{tcolorbox}
    \caption{Axes extraction prompt.}
\label{fig:axes-extraction-prompt}
\end{figure}

\begin{figure}
\centering
\begin{tcolorbox}[colback=gray!5, colframe=gray!80, title=Visualization redrawing prompt, width=\linewidth]
Generate the matplotlib code to generate this \{CHART\_TYPE\}, using the tabular data below.

Provide only the code as output, including the table values represented as a list or a numpy array.
\{TABLE\}
\end{tcolorbox}
    \caption{Visualization redrawing prompt.}
\label{fig:redrawing-prompt}
\end{figure}

\begin{figure}[!ht]
\centering
\begin{tcolorbox}[colback=gray!5, colframe=gray!80, title=QA prompt, width=\linewidth]

\textbf{If table-based QA}: \{TABLE\}\\

\{QUESTION\}\\

\textbf{If MCQ}: Provide the correct answer among the following choices: \{CHOICES\}

\textbf{If rank}: Provide the answer as a Python list.\\

\textbf{If misleader warning}: Be careful, the following design flaw has been identified in the chart: \{MISLEADER DEFINITION\}\\

\textbf{If axes}: Below is a description of the charts axis labels or legend. \{AXIS\}\\

\textbf{If table}: Below is a table containing the values represented in the chart. \{TABLE\}\\

Provide only the final answer to the question.

\end{tcolorbox}
    \caption{QA prompt.}
\label{fig:qa-prompt}
\end{figure}

\section{Experiment 1 - Additional results}
\label{sec:additional_results}

\paragraph{QA accuracy with confidence intervals}

Table \ref{tab:results-with-ci} presents QA accuracy on the misleading and non-misleading datasets, with confidence intervals.

\begin{table}[!ht]
    \centering
    \resizebox{\linewidth}{!}{
    \begin{tabular}{lcccc}
    \toprule
    &  \multicolumn{2}{c}{Misleading visualizations} & \multicolumn{2}{c}{Non-misleading visualizations} \\
    Model & QA accuracy & CI & QA accuracy & CI \\
    \midrule
    llava-7B & 25.2 & (18.2, 32.2) & 42.7 & (33.9, 50.8) \\
    llava-13B & 25.9 & (18.9, 32.9) & 41.1 & (32.3, 50.0) \\
    qwen2vl-2B & 27.3 & (20.3, 34.3) & 39.5 & (30.7, 48.4) \\
    tinychart-3B & 18.9 & (12.6, 25.9) & 22.6 & (15.3, 29.8) \\
    gpt4 & 30.8 & (23.1, 38.5) & 52.4 & (43.6, 61.3) \\
    chartinstruction-13B & 28.0 & (21.0, 35.7) & 35.5 & (26.6, 43.6) \\
    internvl2.5-2B & 16.1 & (10.5, 22.4) & 44.4 & (35.5, 53.2) \\
    chartgemma-3B & 18.9 & (12.6, 25.2) & 23.4 & (16.1, 31.5) \\
    ovis1.6-9B & 24.5 & (17.5, 31.5) & 57.3 & (48.4, 66.1) \\
    qwen2vl-7B & 22.4 & (15.4, 29.4) & 50.8 & (41.9, 59.7) \\
    internvl2.5-4B & 21.0 & (14.7, 28.0) & 50.0 & (41.1, 58.9) \\
    internvl2.5-8B & 24.5 & (17.5, 31.5) & 57.3  & (48.4, 66.1) \\
    ovis1.6-27B & 26.6 & (19.6, 34.3) & 55.7 & (46.8, 64.5) \\
    gemini-1.5-flash & 28.0 & (21.0, 35.7) & 59.7 & (50.8, 68.6) \\
    gpt4o & 36.4 & (28.7, 44.1) & 64.5 & (55.7, 72.6) \\
    gemini-1.5-pro & 34.3 & (26.6, 42.0) & 66.9 & (58.1, 75.0) \\
    internvl2.5-26B & 21.7 & (15.4, 28.7) & 55.7 & (46.8, 64.5) \\
    internvl2.5-38B & 31.5 & (23.8, 39.2) & 63.7 & (54.8, 72.6) \\
    claude-3.5-sonnet & 39.9 & (31.5, 48.3) & 71.8 & (63.7, 79.8) \\
         \bottomrule
    \end{tabular}}
    \caption{Experiment 1 main results with bootstrapped confidence intervals (n=5000) (\%). }
    \label{tab:results-with-ci}
\end{table}

\paragraph{QA accuracy by question type} Table \ref{tab:question_type} provides the QA accuracy per question type for each MLLM and the random baseline.

10 MLLMs perform worse than or equal to the random baseline on MCQs for misleading visualizations. There are only two for the non-misleading ones. Free text is the only question type where some MLLMs perform better on misleading instances than on non-misleading ones. We attribute this to the difficulty of the free text questions in the non-misleading VLAT dataset.
While only four MLLMs achieve non-zero accuracy on rank questions with misleading instances, 13 do so for non-misleading ones, indicating both the difficulty of the task and a strong negative impact of the \textit{3D effect} misleader, which affects all rank questions.

\begin{table*}[!ht]
    \centering
    \resizebox{\linewidth}{!}{
    \begin{tabular}{lcccccc}
    \toprule
    & \multicolumn{2}{c}{MCQ (n=115)} & \multicolumn{2}{c}{Free-text (n=20)} & \multicolumn{2}{c}{Rank (n=8)} \\
    Model & Misleading & Non-misleading & Misleading & Non-misleading & Misleading & Non-misleading \\
    \midrule
    Random baseline & 31.3 & 33.9  &  0.0  &  0.0  &   0.1 & 0.1 \\
  llava-7B & 31.3 & 54.2 & 0.0 & 5.0 & 0.0 & 0.0 \\
    llava-13B & 31.3 & 53.1 & 5.0 & 0.0 & 0.0 & 0.0 \\
    qwen2vl-2B & 33.0 & 51.0 & 5.0 & 0.0 & 0.0 & 0.0 \\
    tinychart-3B & 22.6 & 29.2 & 5.0 & 0.0 & 0.0 & 0.0 \\
    gpt4 & 36.5 & 64.6 & 0.0 & 5.0 & \textbf{25.0} & 25.0 \\
    chartinstruction-13B & 33.9 & 44.8 & 5.0 & 5.0 & 0.0 & 0.0 \\
    internvl2.5-2B & 19.1 & 56.2 & 5.0 & 5.0 & 0.0 & 0.0 \\
    chartgemma-3B & 22.6 & 29.2 & 5.0 & 5.0 & 0.0 & 0.0 \\
    ovis1.6-9B & 26.1 & 69.8 & 25.0 & 15.0 & 0.0 & 12.5 \\
    qwen2vl-7B & 27.0 & 62.5 & 5.0 & 0.0 & 0.0 & 37.5 \\
    internvl2.5-4B & 24.4 & 61.5 & 5.0 & 10.0 & 12.5 & 12.5 \\
    internvl2.5-8B & 27.8 & 69.8 & 10.0 & 15.0 & 12.5 & 12.5 \\
    ovis1.6-27B & 32.2 & 60.4 & 5.0 & 40.0 & 0.0 & 37.5 \\
    gemini-1.5-flash & 33.0 & 70.8 & 10.0 & 10.0 & 0.0 & \textbf{50.0} \\
    gpt4o & 40.9 & 72.9 & 25.0 & 30.0 & 0.0 & 50.0 \\
    gemini-1.5-pro & 37.4 & 72.9 & 30.0 & \textbf{45.0} & 0.0 & \textbf{50.0} \\
    internvl2.5-26B & 24.4 & 67.7 & 15.0 & 20.0 & 0.0 & 0.0 \\
    internvl2.5-38B & 33.0 & 75.0 & \textbf{35.0} & 25.0 & 0.0 & 25.0 \\
    claude-3.5-sonnet & \textbf{44.4} & \textbf{85.4} & 25.0 & 25.0 & 12.5 & 25.0 \\
    \bottomrule
    \end{tabular}}
    \caption{Experiment 1 results by question type (\%), with their number of occurence (\textit{n}). The best results are marked in bold.}
    \label{tab:question_type}
\end{table*}

\begin{figure}
    \centering
    \includegraphics[width=\linewidth]{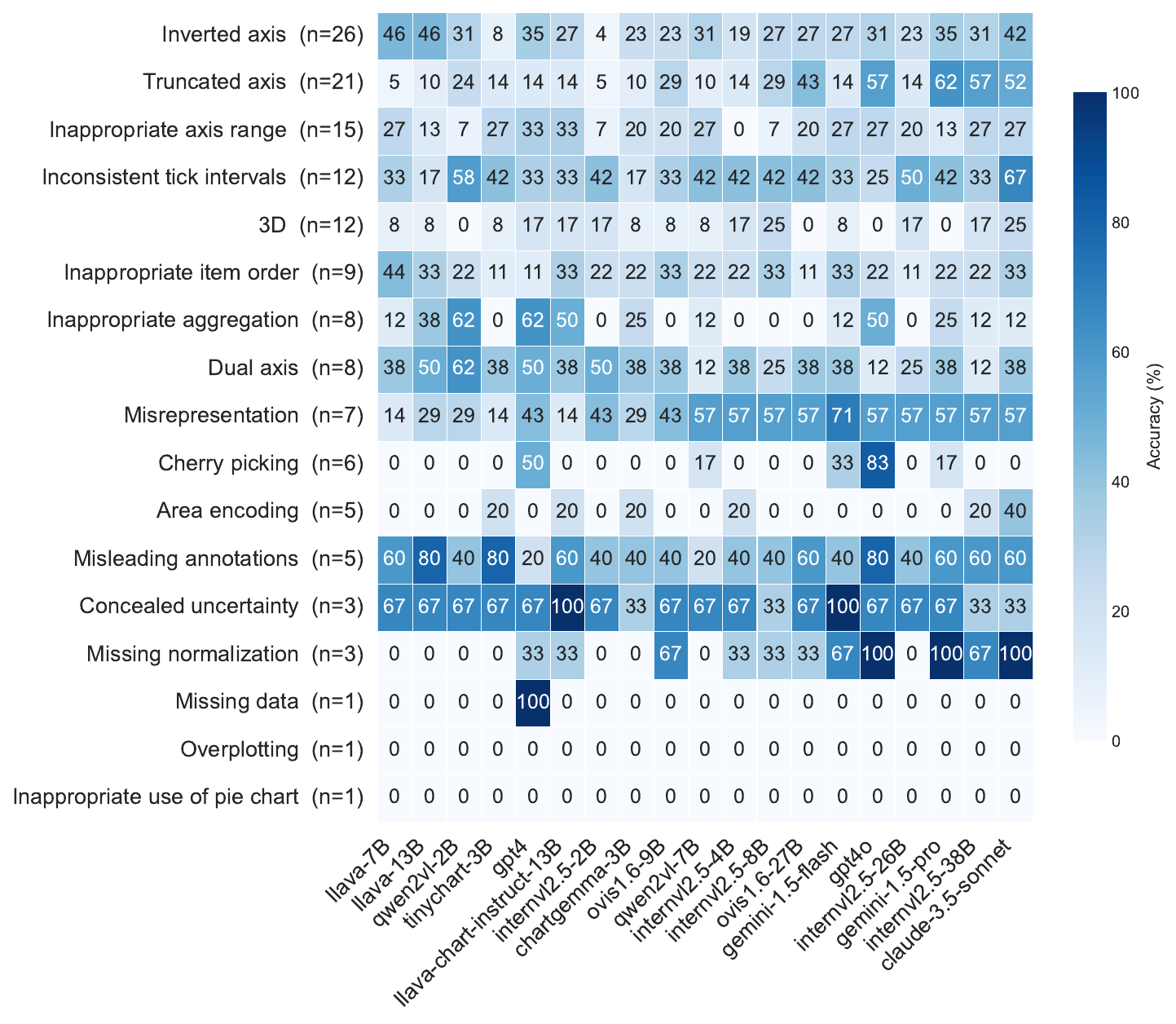}
    \caption{QA Accuracy (\%) per type of misleader.}
    \label{fig7}
\end{figure}

\paragraph{QA accuracy by misleader}  Figure \ref{fig7} provides the QA accuracy per misleader for each MLLM.

\paragraph{QA accuracy on CHARTOM}

Table \ref{tab:chartom} compares the QA accuracy on the misleading and non-misleading subsets of CHARTOM, which are based on the same underlying data tables.
The change ($\Delta$) in accuracy between the two subsets is larger than 20 pp for 14 out of 19 models. The exceptions are datasets on which the MLLMs achieve accuracy below 40\% for non-misleading visualizations, indicating limited chart understanding abilities. Improved performance on the non-misleading subset does not consistently result in a reduced $\Delta$. These results further indicate that improving MLLMs' general chart understanding does not entirely address their vulnerability to misleading visualizations.

\begin{table}[!ht]
    \centering
    \resizebox{\linewidth}{!}{
    \begin{tabular}{lccc}
    \toprule
    Model & Misleading CHARTOM & Non-misleading CHARTOM & $\Delta$\\
    \midrule
    llava-7B & 19.6  & 39.3 & 19.7 \\
    llava-13B & 16.1 & 32.1 & 16.0 \\
    qwen2vl-2B & 21.4 & 44.6 & 23.2 \\
    tinychart-3B & 16.1 & 25.0 & 8.9 \\
    gpt4 & 25.0 & 46.4 & 21.4 \\
    chartinstruction-13B & 23.2 & 37.5 & 14.3 \\
    internvl2.5-2B & 16.1 & 44.6 & 28.5 \\
    chartgemma-3B & 14.3 & 17.9 & 3.6 \\
    ovis1.6-9B & 23.2 & 53.6 & 30.4 \\
    qwen2vl-7B & 17.9 & 53.6 & 35.7 \\
    internvl2.5-4B & 21.4 & 50.0 & 28.6 \\
    internvl2.5-8B & 25.0 & 53.6  & 28.6 \\
    ovis1.6-27B & 17.9 & 51.8 & 33.9 \\
    gemini-1.5-flash & 19.6 & 53.6 & 34.0 \\
    gpt4o & 23.2 & 64.3 & 41.1 \\
    gemini-1.5-pro & 28.6 & 64.3 & 35.7 \\
    internvl2.5-26B & 21.4 & 53.6 & 32.2 \\
    internvl2.5-38B & 28.6 & 62.5 & 33.9 \\
    claude-3.5-sonnet & 37.5 & 62.5 & 25.0\\
         \bottomrule
    \end{tabular}}
    \caption{QA accuracy by subset on CHARTOM (\%).}
    \label{tab:chartom}
\end{table}

\paragraph{Results with models released in 2025}

While the focus of this work is on the MLLMs released in 2023-2024, we provide preliminary results with three commercial models released in 2025: GPT4.1 \citep{openai2023gpt4}, GPT5-mini, and Gemini-2.5-flash-lite \citep{google2025gemini}. Their QA accuracy is reported against the best model from 2024, Claude-3.5-sonnet, in Table \ref{tab:2025-models}. All 2025 models outperform Claude-3.5-sonnet on misleading instances. GPT5-mini even achieves an accuracy above 50\%. However, these MLLMs also perform better on the non-misleading dataset, and a large change ($\Delta$) in accuracy of more than 20\% remains for all of them.

\begin{table}
    \centering
    \resizebox{\linewidth}{!}{
    \begin{tabular}{lccc}
    \toprule
    Model & Misleading visualizations & Non-misleading visualizations &  ChartQA\\
    \midrule
    Claude-3.5-sonnet & 39.9  & 71.8 & \textbf{90.8} \\
    \midrule
    Gemini-2.5-flash & 42.0 &  66.9 & 76.8 \\
    GPT5-mini & 53.2 & 81.5 & 88.2\\ 
    GPT5 & \textbf{55.9}  & \textbf{89.5}   &  89.6 \\

         \bottomrule
    \end{tabular}}
    \caption{ QA accuracy of Claude-3.5-sonnet compared with recent models released in 2025 (\%). }
    \label{tab:2025-models}
\end{table}

\section{Experiment 1 - Impact of parametric knowledge}
\label{sec:real-world-knowledge}

We observed in Experiment 1 that Claude-3.5-sonnet, GPT4o, and Gemini-1.5-pro primarily outperform other MLLMs on the real-world subset of the misleading visualizations. We assume this is due to their parametric knowledge, which allows them to answer the question without considering the visualization.
To support this assumption, we identified a subset of 22 real-world MCQS out of 42 that could be answered using world knowledge. We rephrased the questions slightly to make them self-contained, without direct references to the visualization's content. The accuracy of GPT4o on this subset is 77\%. If we provide only the MCQ to GPT4o, without the visualization's image, the accuracy is still 50\%, highlighting indeed a moderate ability to answer the real-world MCQs based on parametric knowledge. In such cases, parametric knowledge effectively serves as a form of protection against real-world misleaders.  However, this finding should be interpreted with caution. In practical scenarios, the underlying data will often be recent and unlikely to be covered by the parametric knowledge. Other factors might explain the performance on the real-world subset, including differences in visualization complexity and question difficulty compared to CALVI and CHARTOM.

\section{Experiment 3 - Additional results}
\label{sec:additional_results_exp3}

\paragraph{QA accuracy by correction method} Table \ref{tab:res_exp3} provides the detailed QA accuracy results for the default prompt and for the six correction methods.

\begin{table}
    \centering
    \resizebox{\linewidth}{!}{
    \begin{tabular}{llccc}
    \toprule
    &  & Misleading visualizations & Non-misleading visualizations\\
    \midrule  
    & Default & 22.4   &  48.4 \\
    & Misleader waring & 23.1 & \textbf{50.8} \\
    & Axes & 21.7 &  48.4\\
    qwen2vl-7B  & Table & 28.0 & 50.0 \\
    & Table + axes & 28.0 & 47.6 \\
    & Table-based QA & \textbf{42.0} & 41.1 \\
    & Redrawn visualization & 32.2 &46.0  \\
    \midrule  
    & Default & 24.5   & \textbf{56.5}   \\
    & Misleader waring & 29.4 & 55.7 \\
    & Axes & 28.0 & 51.6 \\
    ovis1.6-9B & Table & 21.0 & 46.8 \\
    & Table + axes & 23.8 & 49.2 \\
    & Table-based QA & \textbf{39.9} & 47.6 \\
    & Redrawn visualization & 39.2 & 53.2 \\
    \midrule  
    & Default & 23.8   & \textbf{57.3}  \\
    & Misleader waring & 24.5 &  \textbf{57.3}\\
    & Axes & 23.8 & 56.5 \\
    internvl2.5-8B & Table & 25.9 & 50.8 \\
    & Table + axes & 25.2  & 54.8 \\
    & Table-based QA & \textbf{44.1} & 52.4  \\
    & Redrawn visualization & 25.9 & 52.4 \\

         \bottomrule
    \end{tabular}}
    \caption{QA accuracy results for Experiment 3 (\%). }
    \label{tab:res_exp3}
\end{table}

\paragraph{Results with ChartGemma and GPT5-mini}

We apply the two most effective correction methods, table-based QA and redrawing the visualization, to the strongest chart-specialized model, ChartGemma, and a strong commercial MLLM released in 2025, GPT5-mini.
Table \ref{tab:exp3-chartgemma-GPT5} contains the results.

ChartGemma excels at table extraction and performed very poorly in Experiment 1. This results in a high  $\Delta$ accuracy for table-based QA with Qwen2.5-7B, on both datasets. Due to ChartGemma's weaker visual reasoning abilities, redrawing the visualization does not yield significant improvements.

In contrast, GPT5-mini's performance is already strong, as shown in Appendix  \ref{sec:additional_results}. Hence, it does not benefit from providing the extracted table to Qwen2.5-7B, which has weaker reasoning abilities. This results in a large performance drop on the non-misleading dataset. Redrawing the visualization has no large impact on QA accuracy.

\begin{table}
    \centering
    \resizebox{\linewidth}{!}{
    \begin{tabular}{lcccc}
    \toprule
     & \multicolumn{2}{c}{Misleading visualizations} & \multicolumn{2}{c}{Non-misleading visualizations} \\
    
    Model & Table-based QA & Redrawing & Table-based QA & Redrawing \\
    \midrule
    ChartGemma-3B & +27.3 & +2.1 & +23.4 & +1.6 \\
    \midrule
    GPT5-mini & - 9.1 & +2.8 & - 20.1 & -3.2\\
    
         \bottomrule
    \end{tabular}}
    \caption{Change ($\Delta$) in accuracy (percentage points) compared to Experiment 1 using different inference-time correction methods with ChartGemma-3B and GPT5-mini.}
    \label{tab:exp3-chartgemma-GPT5}
\end{table}

\paragraph{Table-based QA results with DePlot and Matcha}

\begin{figure*}
    \centering
    \includegraphics[width=\textwidth]{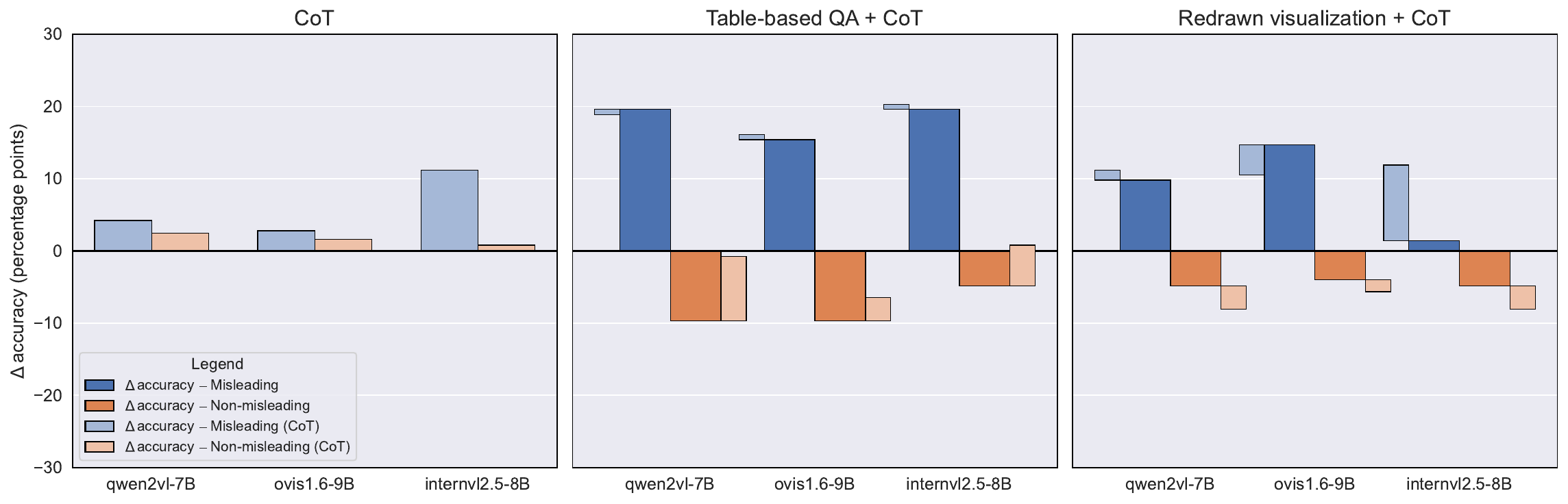}
    \caption{Change ($\Delta$) in accuracy (percentage points) compared to Experiment 1 using Chain-of-Thought (CoT) prompting alone or in combination with correction methods.}
    \label{fig9}
\end{figure*}

\begin{table}
    \centering
    \resizebox{\linewidth}{!}{
    \begin{tabular}{lccc}
    \toprule
    Model & Misleading visualizations & Non-misleading visualizations\\
    \midrule
    DePlot &  \textbf{44.1} &  46.0 \\
    MatCha & 42.7  & 37.9 \\ 
    \midrule
    Qwen2VL-7B & 42.0 & 41.1 \\
    Ovis1.6-9B & 39.9 &  47.6 \\
    InternVL2.5-8B & \textbf{44.1} & \textbf{52.4} \\
    
         \bottomrule
    \end{tabular}}
    \caption{Table-based QA accuracy using Qwen2.5-7B, with different models for table extraction (\%). }
    \label{tab:deplot-matcha}
\end{table}

Table \ref{tab:deplot-matcha} compares the performance of table-based QA using MLLMs with that of two smaller specialized chart-to-table extraction models, DePlot \citep{liu-etal-2023-deplot} and MatCha \citep{liu-etal-2023-matcha}. On misleading visualizations, the specialized models outperform Ovis1.6-9B and Qwen2VL-7B. Only InternVL2.5-8B matches the performance obtained with DePlot. This indicates that smaller models fine-tuned specifically for table extraction constitute a strong alternative to MLLMs for countering misleading visualizations.

However, in non-misleading cases, table-based QA performs worse with DePlot than with InternVL2.5-8B. Moreover, MatCha achieves the lowest performance among all models. We attribute this gap to the high diversity of chart types in the non-misleading dataset, particularly in the VLAT subset, which includes bubble charts and treemaps. Since DePlot and MatCha were not fine-tuned on such chart types, they lack the generalization capabilities of  MLLMs.

\paragraph{Results with Chain-of-Thought prompting}

By default, we evaluate MLLMs in a direct prompting setting where they only output the final answer. Prompting techniques such as Chain-of-Thought (CoT), which requires the MLLM to generate intermediate reasoning steps before providing the final answer, are known to improve performance on several reasoning tasks \citep{10.5555/3600270.3602070}.

Figure \ref{fig9} shows the change in accuracy when using zero-shot CoT prompting, either alone or in combination with one of the two most effective correction methods, table-based QA, and redrawing the visualization. CoT alone improves performance on both misleading and non-misleading visualizations. The gains are generally modest, except for InternVL2.5-8B, where the improvement exceeds 10 pp on misleading visualizations.

Using CoT has no significant impact on misleading visualizations for table-based QA. However, it substantially reduces the negative effects of table-based QA on non-misleading visualizations, bringing the performance drop close to 0 for Qwen2VL-7B and even bringing a net positive change in accuracy for InternVL2.5-8B.

Combining CoT with redrawn visualizations worsens accuracy on non-misleading visualizations. Its effect on misleading visualizations depends on the model. Ovis1.6-9B shows a small decrease, while InternVL2.5-8B experiences a large gain.

Overall, CoT alone is not an effective correction method for misleading visualizations. It can be beneficial when combined with table-based QA, mainly by mitigating the negative impact on non-misleading data. However, its effects are inconsistent across models and settings, and it should therefore be applied with caution.

\begin{figure*}[!ht]
\centering
\includegraphics[width=0.7\linewidth]{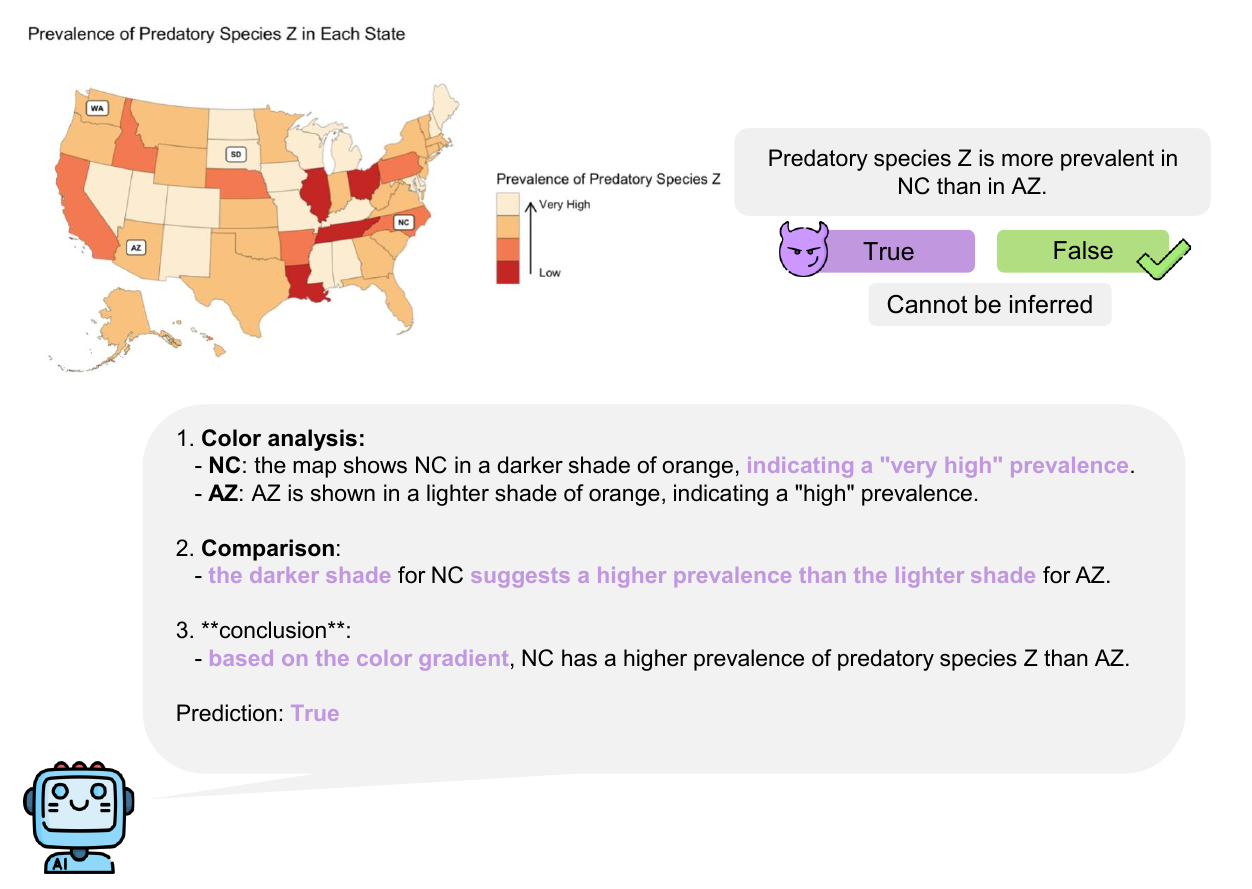}
\caption{Example of CoT reasoning with InternVL2.5-8B on a misleading visualization, where the MLLM fails to answer correctly due to the misleader.}\label{cot1}
\end{figure*}

\begin{figure*}[!ht]
\centering
\includegraphics[width=0.7\linewidth]{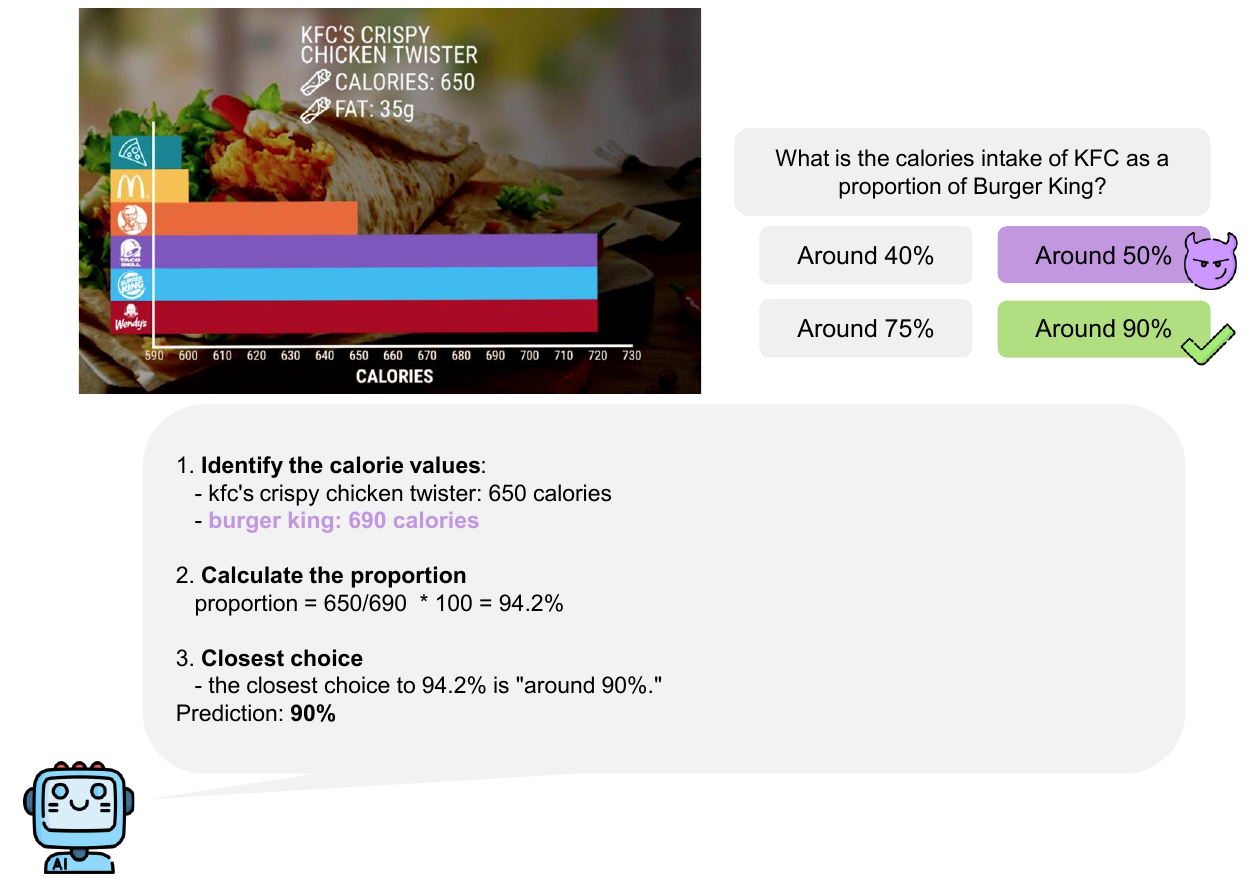}
\caption{Example of CoT reasoning with InternVL2.5-8B on a misleading visualization, where the MLLM answers correctly but the reasoning contains errors.}\label{cot2}
\end{figure*}

\begin{figure*}[!ht]
\centering
\includegraphics[width=0.7\linewidth]{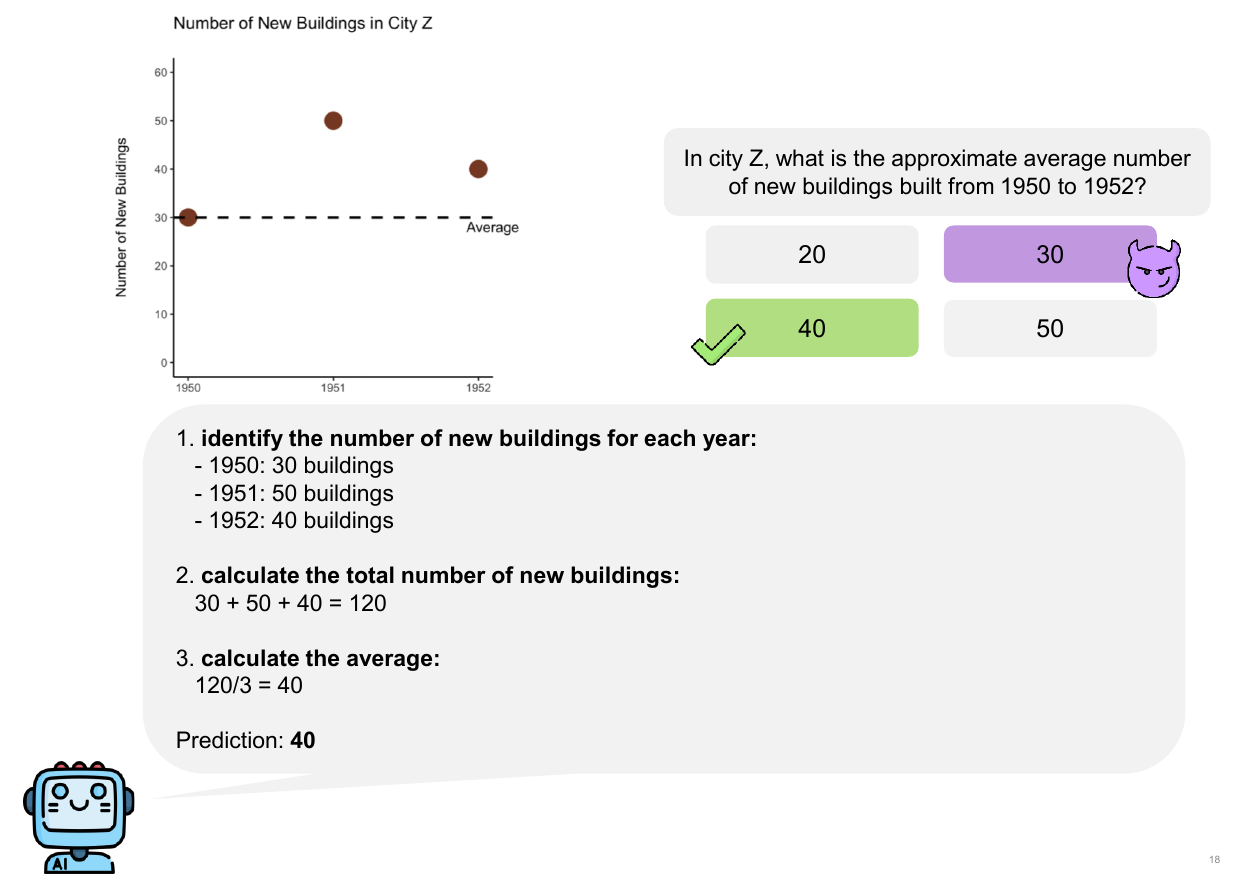}
\caption{Example of CoT reasoning with InternVL2.5-8B on a misleading visualization, where the MLLM answers correctly with a valid reasoning.}\label{cot3}
\end{figure*}

Table \ref{tab:cot-analysis} reports the results of an error analysis of the CoT reasoning chains for a random sample of 30 misleading instances. The majority of the wrong answers and reasoning are due to the presence of the misleader, rather than mathematical mistakes. In a few cases, there is a numerical mistake, but the MLLM answers correctly by rounding to the nearest MCQ choice.
Figure \ref{cot1} shows an instance in which the \textit{inverted} color scale deceives InternVL2.5. Figure \ref{cot2} is an example of a correct answer with errors in the CoT reasoning. In that case, an incorrect value is extracted from the bar chart. Figure \ref{cot3} shows an instance where both the reasoning chain and the answer are correct.

 \begin{table}
    \centering
    \resizebox{\linewidth}{!}{
    \begin{tabular}{lccc}
    \toprule
     & Qwen2VL-7B & Ovis1.6-9B & InternVL2.5-8B \\
    \midrule
Wrong answer due to the misleader & 14 & 18 & 15 \\
Wrong answer due to math reasoning & 4 & 4 & 3 \\
Correct answer, wrong math reasoning & 2 & 2 & 2 \\
Correct answer and reasoning & 10 & 6 & 10 \\
         \bottomrule
    \end{tabular}}
    \caption{Error analysis of the CoT reasoning chains, on a random sample of 30 misleading instances (\%).}
    \label{tab:cot-analysis}
\end{table}

\begin{figure}
\centering
\includegraphics[width=\linewidth]{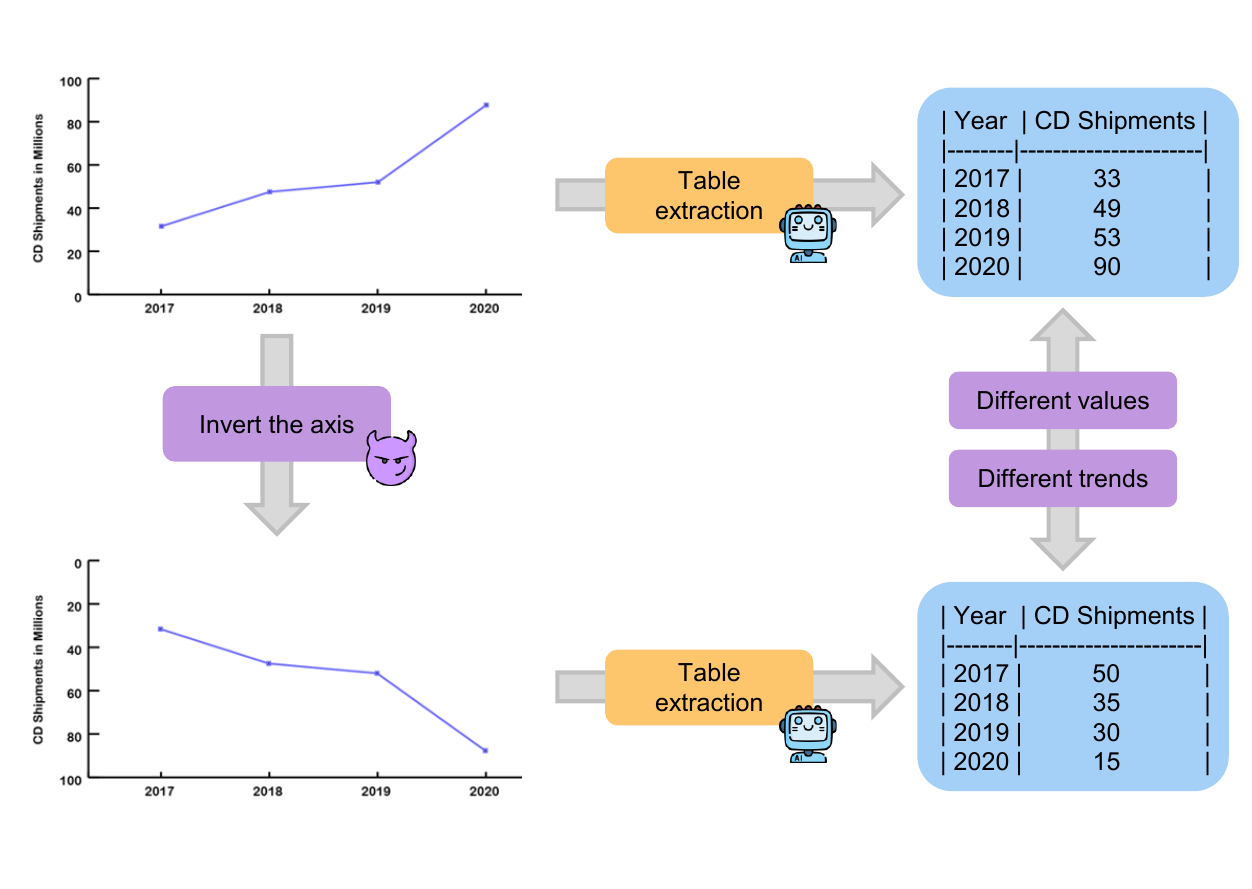}
\caption{Table extraction for a non-misleading visualization and its misleading version with an inverted axis.}\label{fig5}
\end{figure}

\begin{figure}
\centering
\includegraphics[width=\linewidth]{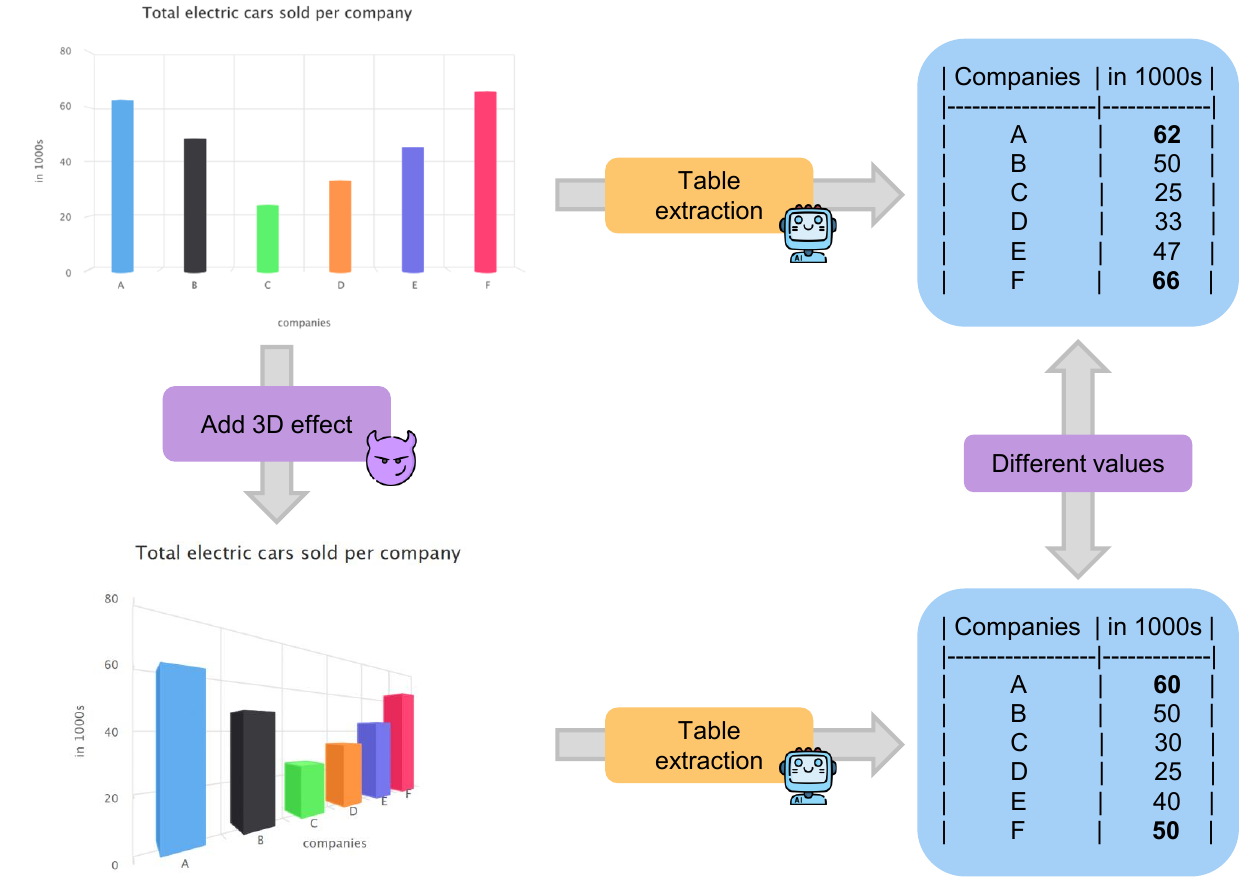}
\caption{Table extraction for a non-misleading visualization and its misleading version with 3D effects.}\label{fig10}
\end{figure}

\begin{figure}
\centering
\includegraphics[width=\linewidth]{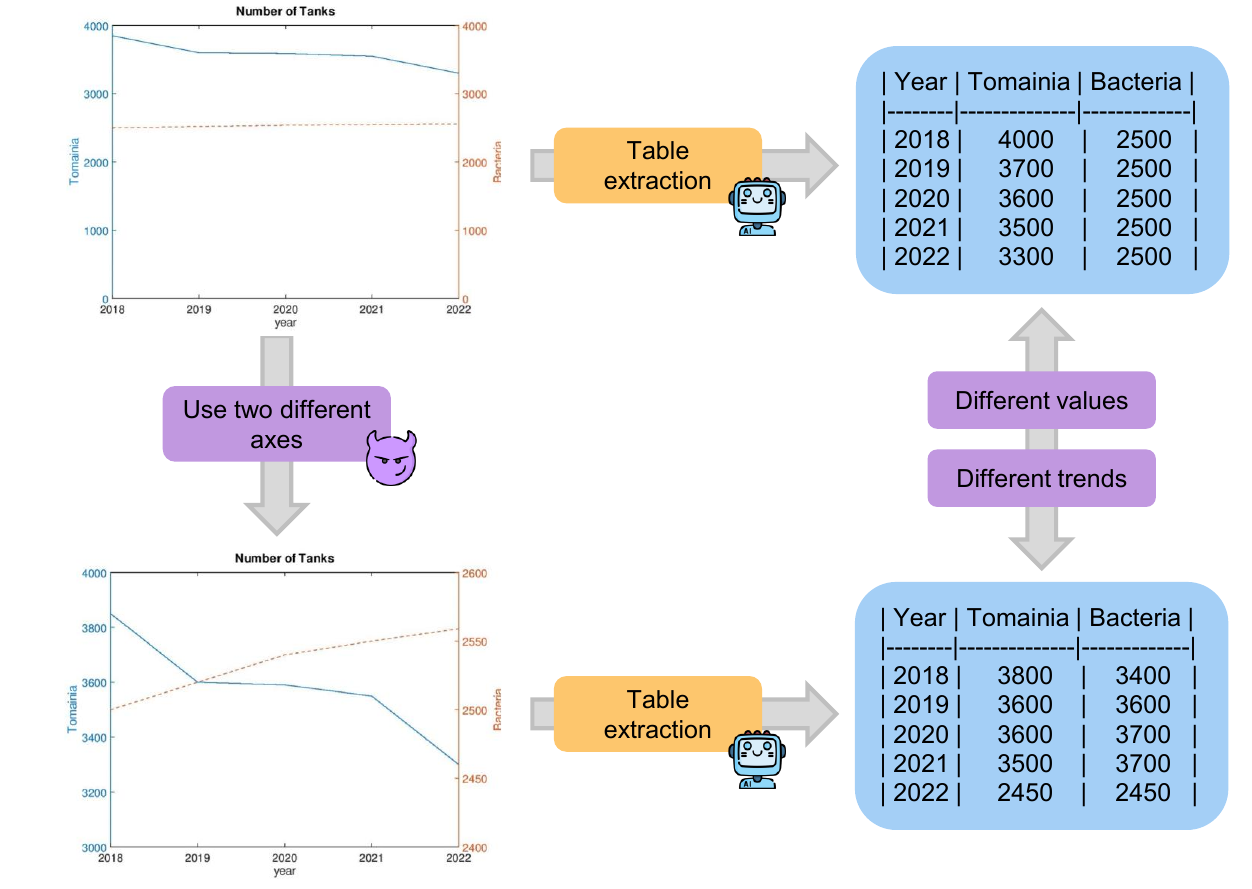}
\caption{Table extraction for a non-misleading visualization and its misleading version with dual axis.}\label{fig11}
\end{figure}

\section{Experiment 3 - Analysis of table extraction}
\label{secA3}

\paragraph{Error examples} Figures \ref{fig5}, \ref{fig10}, and \ref{fig11} illustrate representative examples of table extraction errors made by InternVL2.5-8B due to misleaders. The examples are taken from CHARTOM, where each data table is associated with two instances, one misleading and one non-misleading. In all three cases, the table extracted from the non-misleading instance is accurate. However, the MLLM fails to recover the same table from the misleading version.

In Figure \ref{fig5}, the MLLM fails to correctly align the dots with the inverted axis. In Figure \ref{fig10}, the 3D effects distort the perceived size of the bars on the right side, leading to incorrect value extraction. In Figure \ref{fig11}, the MLLM first extracts values for both lines using the left axis, and then switches to the right axis for 2022, failing to recognize that each line is associated with a different axis.

\paragraph{Impact of table extraction on table-based QA}

We conduct a manual analysis on a random sample of 30 misleading visualizations to assess the impact of the table extraction step on the accuracy of table-based QA. We analyze the three MLLMs from Experiment 3 as well as DePlot, introduced in Appendix \ref{sec:additional_results_exp3}. Table \ref{tab:table-qa analysis} provides the results. In most cases, both steps are incorrect. For all models except DePlot, the second-most-frequent outcome is that both steps are correct.

In some instances, the extracted table is incorrect, but the QA output is correct. This typically occurs when the correct answer is ``not enough information'' and is selected not because of correct reasoning but because the extracted table contains missing or incorrect values.

In other cases, the table extraction is correct, but the QA is incorrect. Except for Ovis1.6-9B, this category is the least prevalent. This occurs mostly with misleaders that are not removed when the visualization is converted to a table. For example, if the misleader is \textit{inappropriate item order}, the entries in the table remain shuffled, which is deceiving. If the misleader is \textit{cherry-picking}, there is still not enough data to answer the question properly.

 \begin{table}
    \centering
    \resizebox{\linewidth}{!}{
    \begin{tabular}{llcc}
    \toprule
    & &  \multicolumn{2}{c}{\textbf{Table-based QA (Qwen2.5-7B)}}  \\
     \multicolumn{2}{c}{\textbf{Table extraction}}  &   Correct & Incorrect \\
    \midrule
       Qwen2VL-7B   &  Correct  & 23.3 &   10.0 \\
            &  Incorrect & 16.7  &   50.0  \\ 
    \midrule
       Ovis1.6-9B   &  Correct  & 26.7 &  20.0 \\
            &  Incorrect &  20.0 &  33.3  \\ 
    \midrule
       InternVL2.5-8B   &  Correct  & 33.3  & 20.0 \\
            &  Incorrect &  10 &   36.7   \\ 
    \midrule
       DePlot   &  Correct  & 13.3 & 10.0 \\
           &  Incorrect & 20.0  &  56.7    \\ 
         \bottomrule
    \end{tabular}}
    \caption{Manual analysis of the impact of table extraction accuracy on table-based QA accuracy, on a random sample of 30 misleading visualizations (\%).}
    \label{tab:table-qa analysis}
\end{table}

 \begin{table}[!ht]
    \centering
    \resizebox{\linewidth}{!}{
    \begin{tabular}{lc}
    \toprule
    Misleader & Table extraction accuracy  \\
    \midrule
    Inverted axis & 15.4 \\
    Truncated axis & 81.0 \\
    Inappropriate axis range & 46.7 \\
    Inconsistent tick intervals & 50.0 \\
    3D & 8.3 \\
    Inappropriate item order & 33.3 \\
    Inappropriate aggregation & 37.5 \\
    Dual axis & 0.0 \\
    Misrepresentation &  85.7 \\
    Cherry-picking & 66.7 \\
    
         \bottomrule
    \end{tabular}}
    \caption{Manual analysis of table extraction accuracy per misleader category, using Internvl2.5-8B (\%).}
    \label{tab:table-extraction-misleader}
\end{table}

\paragraph{Table extraction accuracy by misleader}

Table \ref{tab:table-extraction-misleader} reports our manual analysis of table extraction accuracy with InternVL2.5-8B, the strongest MLLM for table-based QA. We analyze all instances of the ten most frequent misleaders in the misleading visualization dataset. The results show that extraction accuracy varies across misleaders.

For some misleaders, such as truncated axis and misrepresentation, the extraction accuracy exceeds 80\%. The distortions introduced by these misleaders have a limited impact on table extraction. For truncated axis, it is sufficient to align the top of the bars with the vertical axis to recover the values. For misrepresentation, the model often succeeds because the relevant value is explicitly displayed on the bar or pie slice.

In contrast, several misleaders have a severe negative impact on table extraction. When a dual axis is present, all extracted tables are incorrect, as illustrated in Figure \ref{fig11}. Inverted axis and 3D effects also lead to very low extraction accuracy.

Overall, these results show that table-based QA is not a universal correction method for all misleaders. Future work should either improve table extraction for challenging misleaders, such as inverted and dual axis, by creating large-scale synthetic data  and fine-tuning specialized models like DePlot, or by designing dedicated correction methods for these misleaders.

\paragraph{Consistency of table extraction on CHARTOM}
 We conducted a manual analysis of the table extraction outputs of all three MLLMs on CHARTOM, which pairs each question with two visualizations, one misleading and one non-misleading, generated from the same underlying data. For all MLLMs, the extracted tables match exactly in only 4 out of 56 pairs (7.1\%) and partially in 13 to 14 other pairs (23 to 25\%). This further shows the negative impact of misleaders on table extraction accuracy.

\section{Experiment 3 - Analysis of visualization redrawing}

 \begin{table}
    \centering
    \resizebox{\linewidth}{!}{
    \begin{tabular}{lccc}
    \toprule
     & Qwen2VL-7B & Ovis1.6-9B & InternVL2.5-8B \\
    \midrule
    No redrawing for maps & 4 & 4 & 4 \\
    Code did not compile & 4 & 0 &  2  \\
    Incorrect chart design & 4 & 3 & 2  \\
    Correct chart, incorrect values & 7 & 12 &   8  \\
    Correct chart and values & 11 & 11 &  13 \\
         \bottomrule
    \end{tabular}}
    \caption{Error analysis of the generated codes of the redrawn visualizations, on a random sample of 30 misleading instances (\%).}
    \label{tab:code-analysis}
\end{table}

\begin{figure}
\centering
\includegraphics[width=\linewidth]{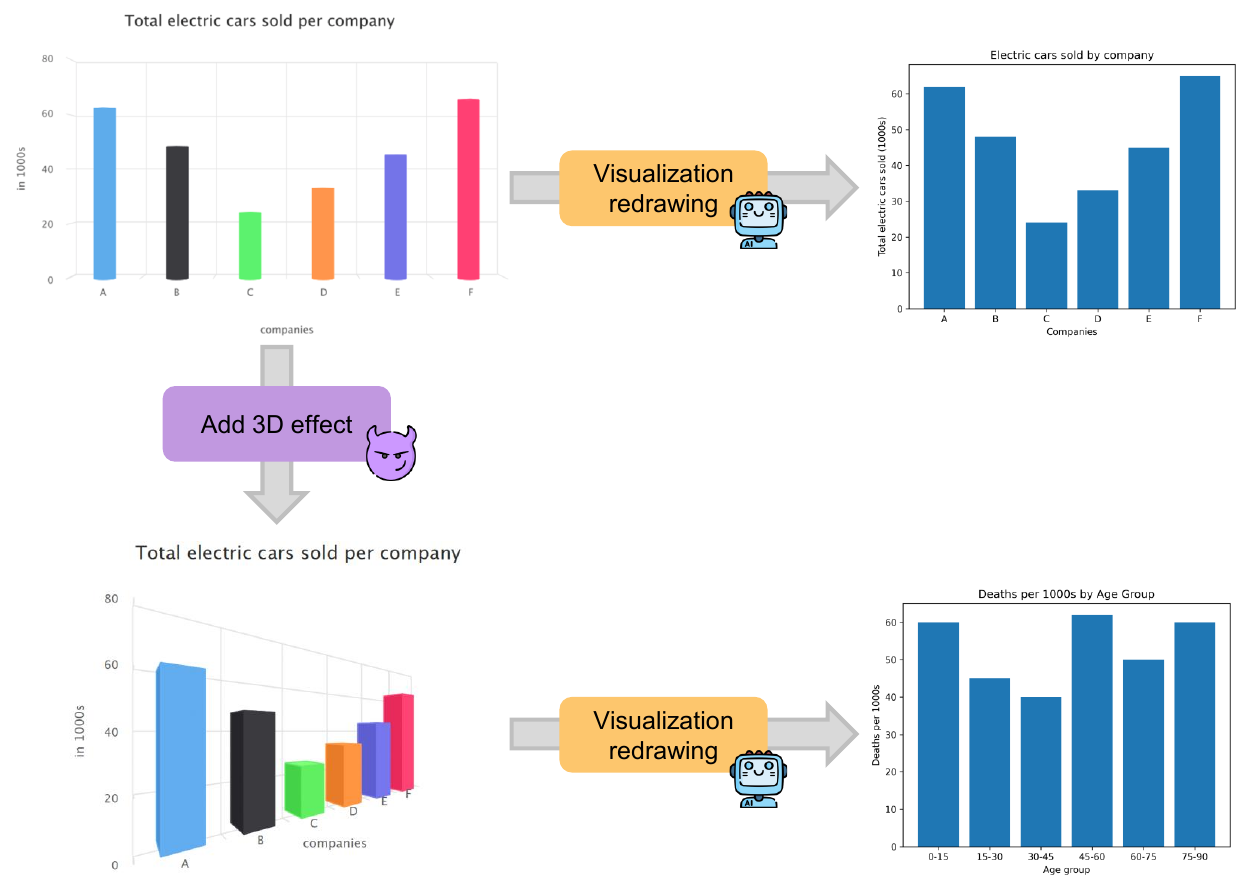}
\caption{Visualization redrawing for a non-misleading visualization and its misleading version with inverted axis.}\label{redraw1}
\end{figure}

Table \ref{tab:cot-analysis} shows the results of a manual analysis of the quality of the redrawn visualizations on a random sample of 30 misleading instances. A correct chart with correct values is drawn in more than 30\% of the cases. The most prevalent issue is redrawing a correct chart type with incorrect values. This is due to error propagation from the table extraction step. In less frequent cases, Qwen2.5-7B generates code that produces the wrong type of visualization or does not compile at all.

\begin{figure}
\centering
\includegraphics[width=\linewidth]{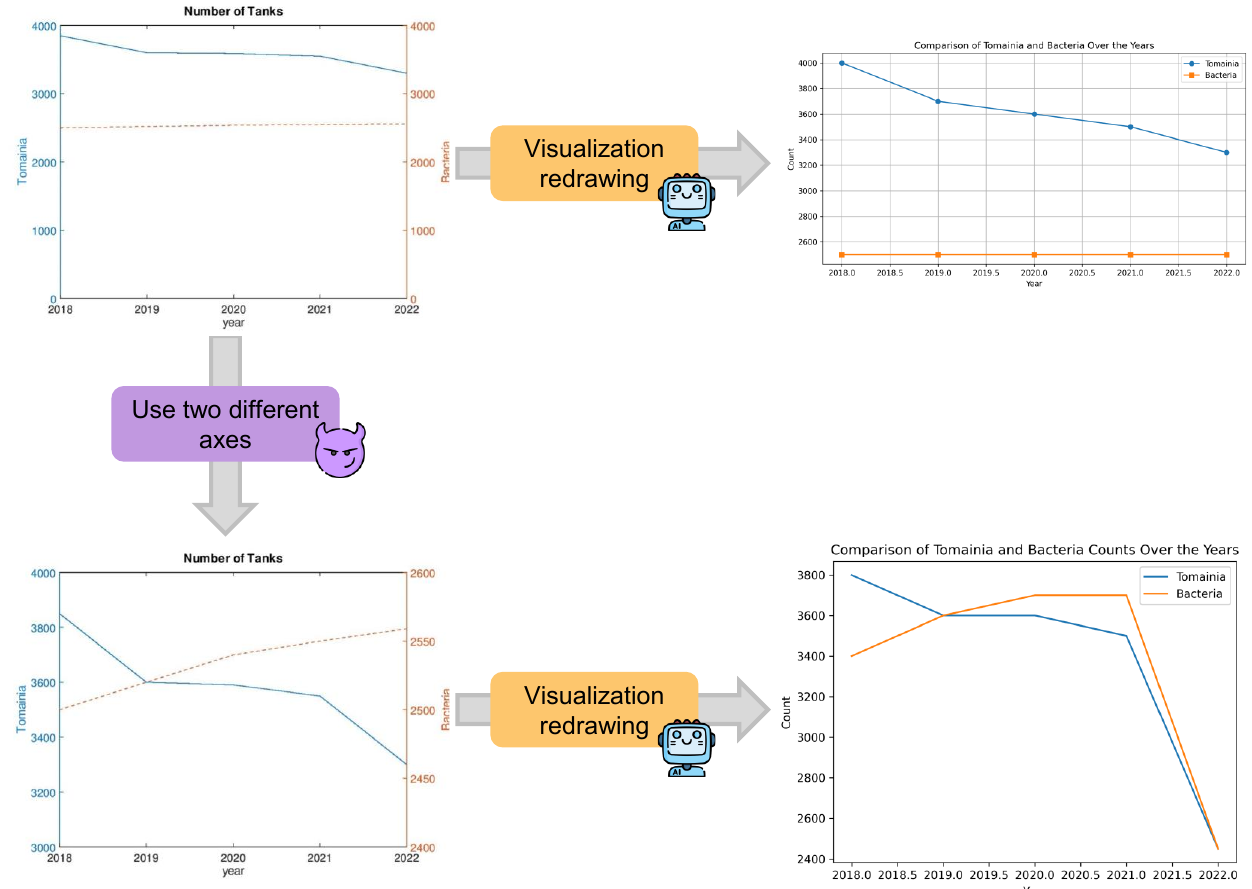}
\caption{Visualization redrawing for a non-misleading visualization and its misleading version with 3D effects.}\label{redraw2}
\end{figure}

\begin{figure}
\centering
\includegraphics[width=\linewidth]{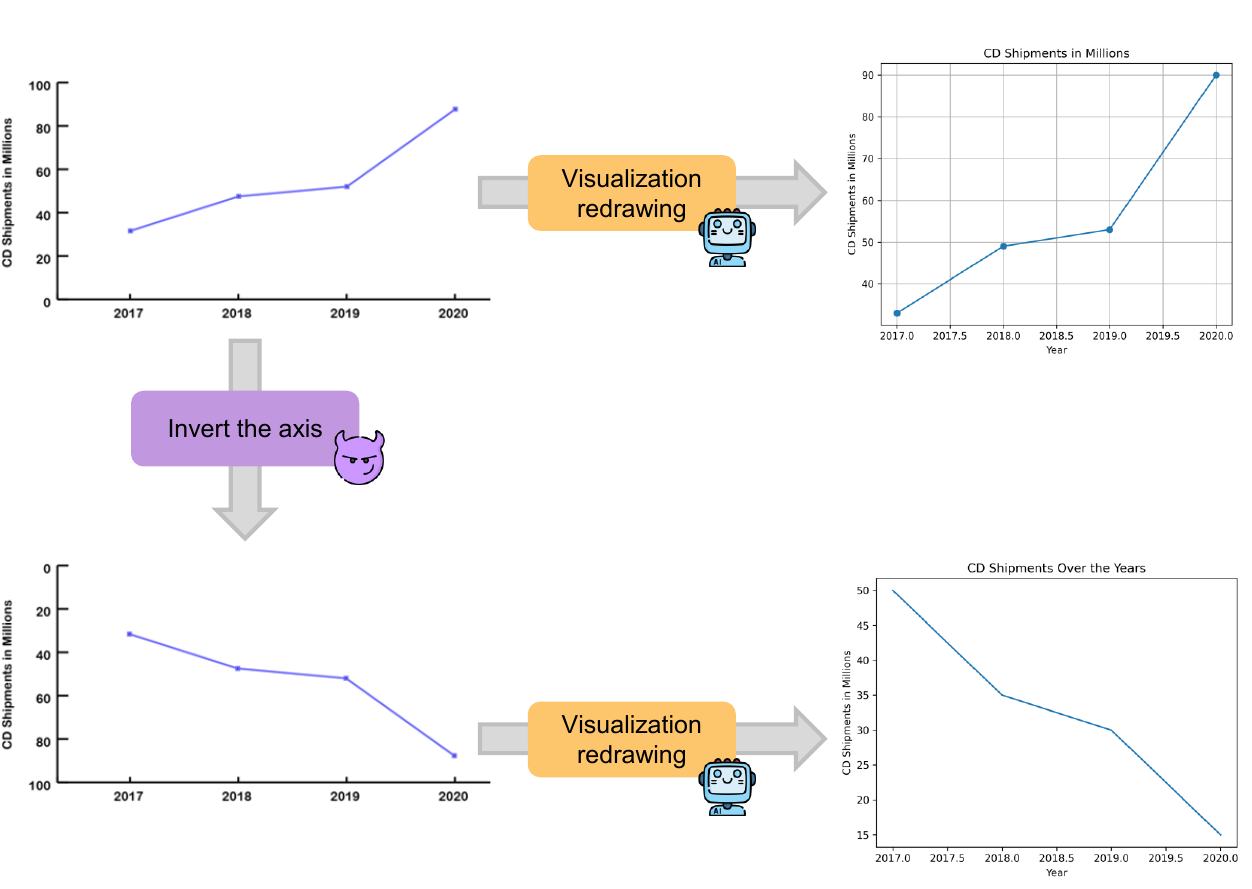}
\caption{Visualization redrawing for a non-misleading visualization and its misleading version with dual axis.}\label{redraw3}
\end{figure}

Figures \ref{redraw1}, \ref{redraw2}, and \ref{redraw3} provide error examples based on pairs of misleading and non-misleading instances of CHARTOM. The examples are the same as those used for table extraction in Figures \ref{fig5}, \ref{fig10}, and \ref{fig11}.

\section{Experiment 3 - Analysis of axes extraction} 
\label{sec:axis_extraction}

 \begin{table}
    \centering
    \resizebox{\linewidth}{!}{
    \begin{tabular}{llcc}
    \toprule
    & &  \multicolumn{2}{c}{\textbf{Visualization + axes QA}}  \\
     \multicolumn{2}{c}{\textbf{Table extraction}}  &   Correct & Incorrect \\
    \midrule
       Qwen2VL-7B   &  Correct  &  16.7 & 50.0  \\
            &  Incorrect &  13.3 &  20.0  \\ 
    \midrule
       Ovis1.6-9B   &  Correct  & 20.0  & 56.7 \\
            &  Incorrect & 10.0 & 13.3   \\ 
    \midrule
       InternVL2.5-8B   &  Correct  & 23.3  & 60.0 \\
            &  Incorrect &  6.7 &   10.0   \\ 
         \bottomrule
    \end{tabular}}
    \caption{Manual analysis of the impact of axes extraction accuracy on visualization+axes QA accuracy, on a random sample of 30 misleading visualizations (\%).}
    \label{tab:axis_analysis}
\end{table}

We conduct a manual analysis on a random sample of 30 misleading visualizations to assess the impact of the axes extraction step on the accuracy of the visualization+axes correction method. The results are reported in Table \ref{tab:axis_analysis}.

Compared to the table extraction results shown in Table \ref{tab:table-qa analysis}, axes extraction is often accurate, with 20 to 25 out of 30 instances containing correct axis information. However, this high extraction accuracy does not translate into strong QA performance. For all MLLMs, at least half of the instances exhibit correct axes but incorrect QA predictions. Two main factors explain this behavior. First, the MLLMs often appear to ignore the extracted axes information and remain primarily influenced by the image modality. Second, the axes are unrelated to the distortions introduced by several misleaders, such as \textit{3D effects} or \textit{misrepresentation}, which limits their corrective value.

Axes extraction errors occur most frequently for \textit{inconsistent tick intervals}, where the MLLMs tend to hallucinate evenly spaced ticks along the axis.

\end{document}